\documentclass{IEEEtaes}

\usepackage{color,array,amsthm}
\usepackage{graphicx}

\usepackage{algorithmicx,algorithm}
\usepackage{algpseudocode}
\usepackage{algorithm,algpseudocode}
\usepackage{makecell,booktabs}
\usepackage{multirow}
\usepackage{multicol}  
\usepackage{bm}
\usepackage{siunitx}
\usepackage{xcolor}
\usepackage{hyperref}
\usepackage{enumitem}
\usepackage{subfig}
\usepackage{xspace}
\usepackage{etoolbox}
\usepackage{pdfcomment}
\usepackage{amsmath}
\usepackage{amssymb}
\usepackage[normalem]{ulem}
\usepackage{marvosym}   
\usepackage{soul}
\hypersetup{hidelinks}

\usepackage{cleveref}
\crefname{section}{Sec.}{Sec.}
\crefname{figure}{Fig.}{Fig.}
\crefname{table}{Table}{Table}
\crefname{algorithm}{Algorithm}{Algorithm}
\crefname{equation}{Eq.}{Eq.}
\crefname{appendix}{\textbf{Appendix}}{\textbf{Appendix}}

\crefformat{subsection}{Sec.~\thesection-#2#1#3}

\newcommand{\multirowoffset}{-0.5\dimexpr \aboverulesep + \belowrulesep + \cmidrulewidth}

\newcommand{\SystemName}{DiffPoL\xspace}

\begin{document}

\title{Learning Satellite Pattern-of-Life Identification: A Diffusion-based Approach}

\author{YONGCHAO YE}
\author{XINTING ZHU}
\affil{Department of Data Science, City University of Hong Kong, Hong Kong SAR, China}

\author{XUEJIN SHEN}
\author{XIAOYU CHEN}
\affil{Chengdu Atom Data Tech Co., Ltd, Chengdu, China}


\author{S. JOE QIN}
\affil{School of Data Science, Lingnan University, Hong Kong SAR, China}

\author{LISHUAI LI}
\affil{Department of Data Science, City University of Hong Kong, Hong Kong SAR, China}



\corresp{Corresponding author: Lishuai Li}

\authoraddress{Yongchao Ye, Xinting Zhu are with the Department of Data Science, City University of Hong Kong, Hong Kong SAR, China
(e-mail: \href{mailto:yongchao.ye@my.cityu.edu.hk}{yongchao.ye@my.cityu.edu.hk}; \href{mailto:xt.zhu@my.cityu.edu.hk}{xt.zhu@my.cityu.edu.hk}).
Xuejin Shen and Xiaoyu Chen are with Chengdu Atom Data Tech Co., Ltd, Chengdu, China
(e-mail: \href{mailto:xuejin.shen@starinnov.com}{xuejin.shen@starinnov.com}; \href{mailto:seanchen@outlook.com}{seanchen@outlook.com}).
S. Joe Qin is with the School of Data Science, Lingnan University, Hong Kong SAR, China
(e-mail: \href{mailto:joeqin@ln.edu.hk}{joeqin@ln.edu.hk}).
Lishuai Li is with the Department of Data Science, City University of Hong Kong, Hong Kong SAR, China
(e-mail: \href{mailto:lishuai.li@cityu.edu.hk}{lishuai.li@cityu.edu.hk}).
}


\supplementary{
This work was supported by the Hong Kong Innovation and Technology Commission Innovation and Technology Fund (Project No. GHP/145/20).}

\markboth{YE ET AL.}{LEARNING SATELLITE PATTERN-OF-LIFE IDENTIFICATION}
\maketitle

\begin{abstract}
As Earth's orbital satellite population grows exponentially, effective space situational awareness becomes critical for collision prevention and sustainable operations. 
Current approaches to monitor satellite behaviors rely on expert knowledge and rule-based systems that scale poorly. 
Among essential monitoring tasks, satellite pattern-of-life (PoL) identification—analyzing behaviors like station-keeping maneuvers and drift operations—remains underdeveloped due to aerospace system complexity, operational variability, and inconsistent ephemerides sources. 
We propose a novel generative approach for satellite PoL identification that significantly eliminates the dependence on expert knowledge.
The proposed approach leverages orbital elements and positional data to enable automatic pattern discovery directly from observations. 
Our implementation uses a diffusion model framework for end-to-end identification without manual refinement or domain expertise. The architecture combines a multivariate time-series encoder to capture hidden representations of satellite positional data with a conditional denoising process to generate accurate PoL classifications. 
Through experiments across diverse real-world satellite operational scenarios, our approach demonstrates superior identification quality and robustness across varying data quality characteristics. 
A case study using actual satellite data confirms the approach's transformative potential for operational behavior pattern identification, enhanced tracking, and space situational awareness.
 
\end{abstract}



\section{Introduction}

With the increasing number of space objects operating in Earth's vicinity, space situational awareness (SSA) technology has become more crucial than ever for tracking and predicting orbital maneuvers, ensuring safe and sustainable space operations. 
According to global statistics on orbiting satellite objects, approximately 26,700 satellites were circling Earth by 2023, marking a 6.8\% increase from the previous year \cite{statista2024}. 
This unprecedented growth, particularly within the heavily utilized Low Earth Orbit (LEO) and Geosynchronous Earth Orbit (GEO) regimes, has significantly elevated the potential for collisions and operational interference, underscoring the pressing need for advanced SSA capabilities.

In SSA, one effective approach is tracking and characterizing on-orbit satellite pattern-of-life (PoL) behaviors \cite{challenge, roberts2023geosynchronous}. 
Satellite PoL refers to the typical behaviors and operational modes of satellites as operators issue commands for different mission types, such as station-keeping \cite{huang2021efficient}, longitudinal-shift maneuvers \cite{roberts2023geosynchronous}, and retirement maneuvers \cite{li2018historical}. 
For instance, station-keeping maneuvers, a primary behavior in geostationary equatorial orbit (GEO) satellite operations, are performed to maintain the satellite's position within a designated orbital slot, counteracting perturbations such as gravitational influences from the sun and moon, and the irregular shape of the Earth \cite{lee1999analysis}. 
Station-keeping maneuvers can be further classified into north-south (NS) and east-west (EW) types, referring to corrections in geographic latitude and longitude, respectively. 

\begin{figure}
    \centering
    \subfloat[Satellite on-orbit maneuvers]{
        \includegraphics[width=0.28\linewidth]{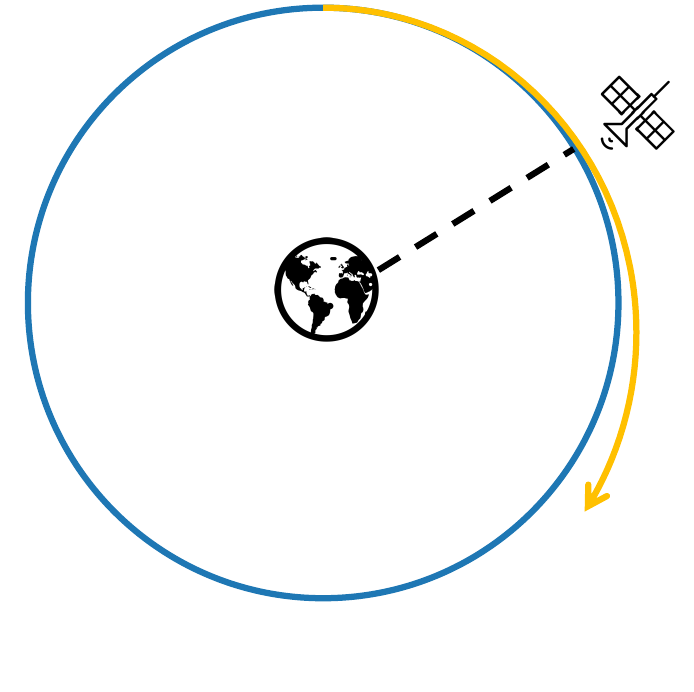}
    }
    \hfill
    \subfloat[Station-keeping maneuvers in NS/EW directions]{
        \includegraphics[width=0.56\linewidth]{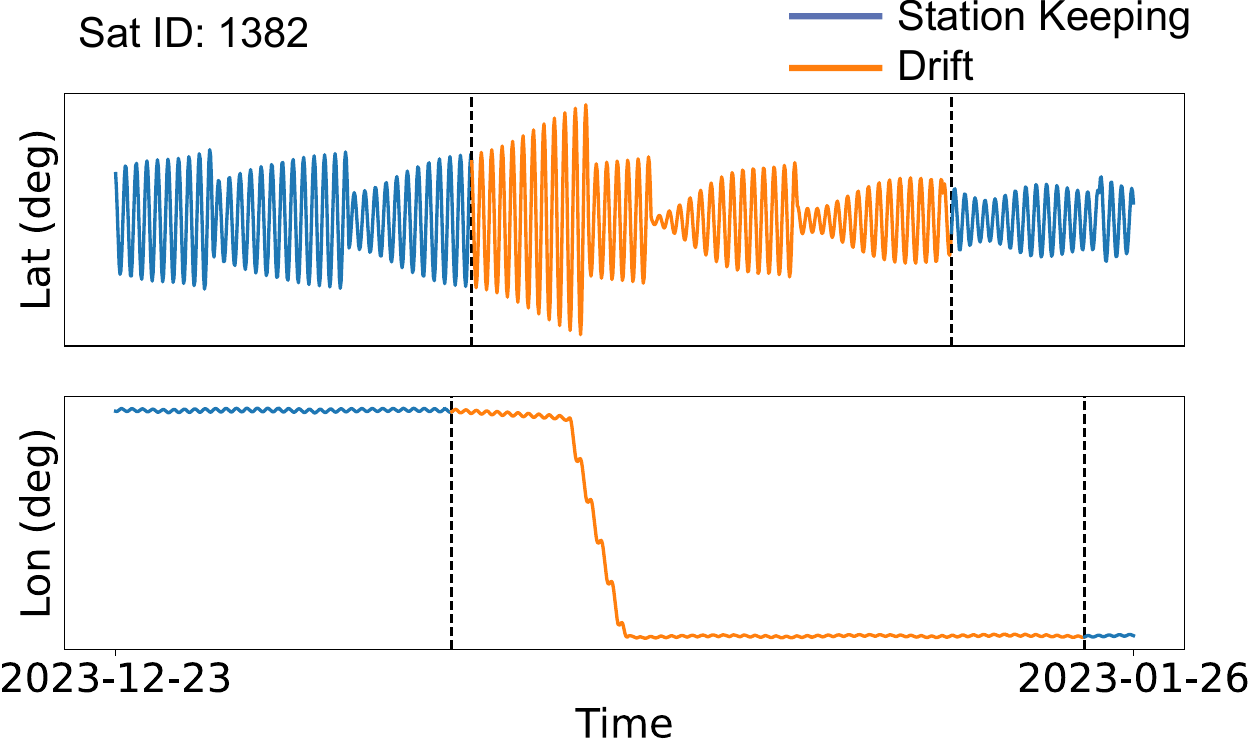}
    }
    \caption{Example of pattern-of-life behavior of a GEO satellite in station-keeping maneuvers.}
    \label{fig:pol}
\end{figure}

\cref{fig:pol} illustrates an example of GEO satellite PoL station-keeping behaviors with (a) illustrating a schematic of Earth-orbiting satellite maneuvers and part (b) showing actual positional data for a satellite, where blue regions indicate active station-keeping and orange regions show drift periods across both latitude and longitude dimensions.
The identification of these behavioral patterns is paramount not only for collision avoidance and improved tracking capabilities but also for detecting anomalous behaviors that might indicate system malfunctions, cybersecurity threats, or unauthorized activities. Furthermore, understanding satellite PoL contributes substantially to enhanced mission planning, more efficient resource allocation, and a more comprehensive space situational awareness framework that fosters a stable and secure operational environment in space.


However, identifying satellite PoL encounters several significant challenges in practice. 
First, due to the complexity of aerospace systems like satellites, commonly adopted rule-based or statistical methods rely heavily on extensive expert knowledge \cite{lemmens2014two, roberts2020satellite, zollo2024comparison}, making them labor-intensive and prone to human overload with the blooming number of daily generated operational data and limiting their scalability with increasing satellites to track and applicability across different satellites and systems. 
Second, the behaviors of different satellites can vary greatly not only because of differing control strategies but also because of diverse propulsion types,posing challenges in modeling and predicting individual on-orbit behavior patterns.
This variability makes it difficult to develop universal rules or heuristics, further justifying the need for adaptive data-driven approaches that can learn these patterns from the data itself rather than relying on pre-programmed rules.
Third, another typical issue in identifying satellite PoL behaviors are the variations in data quality and resolution between different ephemerides sources.
Satellite positional data comes from various sources with differing levels of precision and availability, ranging from high-fidelity ephemerides derived from models like Vector Covariance Message (VCM) that provide detailed orbital information, to the more widely available but less accurate formats such as Two-Line Element (TLE) sets \cite{challengereport}.
The latter offers significantly lower sampling rates and resolution, creating substantial differences in data quality across satellite monitoring systems. 
These differences lead to incomplete or inconsistent series, making it difficult to develop generic PoL identification algorithms that maintains accuracy and reliability across different data sources.
For example, in our another attempt, we found that physical rule-based machine learning methods had good accuracy but showed deployment failures in practice when the observation frequency was inconsistent, which is further discussed in \cref{sec:exper}.

While there have been some preliminary efforts to use supervised learning and deep neural networks for PoL identification \cite{solera2023geosynchronous, roberts2021geosynchronous}, these approaches have been limited in scope and effectiveness. 
The outputs of these models often require further refinement to produce a usable PoL sequence, and they haven't fully addressed the fundamental challenges of end-to-end, data-driven PoL identification. 
Creating a truly autonomous, data-driven approach to satellite PoL identification that eliminates the need for expert intervention represents a significant advancement for the field.
These challenges are also present in similar sequence segmentation problems, such as action \cite{mstcn++, xia2023timestamp} and time-series \cite{utime, yuLaSSID}, where accurately delineating sequences is crucial. 
While these domains have seen advances in data-driven methods, satellite PoL identification presents unique challenges that require novel approaches—specifically its domain-specific nature, varied satellite behavior patterns, and inconsistent data sources.

To address the existing gaps, we propose a data-driven approach, \SystemName for satellite PoL identification that leverages diffusion model. 
This approach effectively learns the complex patterns among various satellites directly from positional data and achieves end-to-end PoL identification without relying on expert-crafted rules.
Inspired by previous studies \cite{mstcn++, ding2024temporal, diffact}, we design this method based on the following primary motivations:
(1) Orbital elements can effectively reflect the status of satellite maneuvers and are sufficiently representative of the position and trajectory of satellites in celestial dynamics, making them ideal inputs for data-driven learning approaches that can automatically discover patterns without manual feature engineering. 
(2) While conventional sequence modeling methods like RNNs or transformers can achieve high point-wise accuracy, their sequence-level performance was unsatisfactory in our experiments. 
The diffusion model's step-by-step denoising process significantly improves temporal-consistent sequence generation—something other methods could not achieve directly. 
(3) The progressive denoising process of diffusion models eliminates the need for manual refinement that is typically required in other approaches, which is labor-intensive and relies heavily on domain knowledge, as seen in other conventional methods.

While the diffusion model shows powerful modeling ability in various sequences modeling tasks \cite{difftraj, wen2023diffstg}, it poses a significant challenge when applied to satellite PoL identification.
Firstly, satellite PoL identification is not a standard conditional generation problem, necessitating the adaptation of the diffusion model.
Secondly, the position sequences of satellites constitute a multivariate time series, requiring the model to simultaneously consider dependencies across both temporal and feature dimensions.
Addressing these challenges is crucial for successfully applying the diffusion model to satellite PoL identification and ensuring accurate, robust outcomes.

Building on the identified motivations and addressing the existing gaps, we propose a diffusion-based method for satellite PoL identification, \SystemName, which resolves the satellite PoL identification problem as a conditional PoL sequence generation task. 
\SystemName incorporates a multivariate time-series encoder that captures the hidden representation of satellite positional observations. 
This representation is subsequently applied as conditional information in the denoising process and fed into a tailored decoder to generate the PoL sequence.
In summary, the contributions of this research are as follows:
\begin{itemize}
    \item We introduce a fully data-driven approach, \SystemName, to satellite PoL identification, marking a paradigm shift from traditional expert-dependent methods toward autonomous, scalable solutions that can adapt to the increasing complexity of the orbital environment.
    To the best of our knowledge, this is the first exploration of satellite PoL identification using the generative model.

    
    \item We validate the effectiveness of \SystemName on real-world satellite datasets and prove its powerful denoising capabilities to achieve accurate end-to-end PoL identification without the need for manual refinement or domain-specific knowledge.
    
    \item We demonstrate the scalability and applicability of \SystemName by comparing it with current practices, and also verify its robustness across different ephemerides sources with varying data quality and resolution, addressing a common challenge in operational satellite behavior surveillance.
    
\end{itemize}

\section{Related Work}\label{sec:related}

\subsection{Satellite PoL identification}
Current practices in satellite PoL identification are still in their developmental stages, heavily relying on extensive expert knowledge \cite{lemmens2014two, li2018historical, roberts2020satellite, zollo2024comparison}. 
For example, Stijn \textit{et al.} detected anomalies in low earth orbit satellite positions through consistency checks between the two-line element (TLE) sets and used background models \cite{lemmens2014two}.
Similarly, Roberts \textit{et al.} utilized longitude to identify geosynchronous satellite maneuvers \cite{roberts2020satellite}.
These traditional approaches, while providing initial capabilities, often prove to be labor-intensive due to the ever-increasing volume of daily operational data generated by the growing number of satellites. 
This inherent limitation in scalability and applicability across diverse satellite types and systems underscores the need for more adaptable techniques

In response to the limitations of traditional methods, researchers have begun to explore the application of supervised learning and deep learning techniques for satellite PoL identification. 
Early efforts in this direction include the use of convolutional neural networks (CNNs) to detect longitudinal shift maneuvers in geosynchronous satellites \cite{roberts2021geosynchronous}. 
Roberts \textit{et al.} also investigated the use of k-means clustering and extended validation techniques for PoL identification \cite{roberts2023geosynchronous, solera2023geosynchronous}.
While these studies represent important early steps in applying machine learning to PoL identification, they often rely on discriminative models like k-means or CNNs, whose outputs may require significant post-processing to ensure temporal consistency. 
This prior experience provides a direct motivation for investigating the potential of data-driven models in addressing the challenges of satellite PoL identification.

\subsection{Time-series segmentation}
Satellite PoL identification can be viewed as a specific instance of the broader problem of time-series segmentation, which involves dividing a sequence of data points over time into meaningful and distinct segments for subsequent analysis \cite{zhang2024self}.
The field of time-series segmentation encompasses a wide variety of methods, ranging from universal techniques applicable across different domains to more specialized approaches tailored for specific data characteristics.
Universal methods are based on unsupervised or self-supervised change point detection \cite{clasp, fluss, deldari21time, wang2024incorporating}.
For example, CLaSP (Classification Score Profile) approach the problem by hierarchically splitting the time series based on the performance of binary classifiers at potential split point \cite{clasp}.
PaTSS (Pattern-based Time Series Semantic Segmentation) focuses on detecting gradual state transitions by learning a distribution over semantic segments within an embedding space derived from frequently occurring sequential pattern \cite{patss}.
However, these methods cannot utilize labels for learning, which limits their performance.

While universal methods offer broad applicability, domain-specific time-series segmentation techniques are tailored to the unique characteristics of data arising from particular fields. 
For example, Perslev \textit{et al.} proposed U-Time, a convolutional neural network based on the U-Net architecture, specifically designed for sleep stage classification using electroencephalography (EEG) data \cite{utime}.
In the domain of human activity recognition, methods have been developed to segment data from wearable sensors using techniques like contrastive learning and order-preserving optimal transport \cite{xia2023timestamp}.
LPTM (Large Pre-trained Time-series Models) introduces adaptive segmentation strategies that are automatically learned during pre-training, allowing the model to effectively process time-series data with varying characteristics from multiple domain \cite{KamarthiP24}.
These methods are customized according to specific data, making them difficult to apply directly to cross-domain tasks like PoL identification.

\subsection{Temporal action segmentation}
In the context of computer vision, temporal action segmentation, tagging each video frame with a corresponding action label, shares similarities with PoL identification in terms of sequence segmentation.
Temporal action segmentation aims to densely identify and label each frame of a video with a corresponding action class.
Several innovative approaches have been developed to address this task, some of which offer valuable insights for satellite PoL identification \cite{ding2024temporal}. 
One notable method is the use of diffusion models for action segmentation, as proposed by Liu \textit{et al.} \cite{diffact}. 
This approach utilizes diffusion models to iteratively generate action predictions from random noise, conditioned on input video features.
A significant focus in temporal action segmentation research is on ensuring temporal consistency in the predicted action sequences \cite{ding2024temporal}.

Multi-stage refinement architectures are a common strategy to improve the temporal coherence and accuracy of action segmentation models. 
For example, MS-TCN (Multi-Stage Temporal Convolutional Network) employs multiple stages of dilated temporal convolutions to capture long-range temporal dependencies and iteratively refine the initial segmentation predictions \cite{mstcn}. 
MS-TCN++ further enhances this architecture with the introduction of dual dilated layers to improve the receptive field of the lower network layers \cite{mstcn++}. 
For skeleton-based action segmentation, HASR (Hierarchical Action Segmentation Refiner) is designed to refine the segmentation results from various backbone models by understanding the overall context of the entire video in a hierarchical manner \cite{ahn2021refining}.
Another prominent direction in ensuring temporal consistency involves the use of unsupervised optimal transport \cite{xu2024temporally, kumar2022unsupervised}. 
ASOT (Action Segmentation Optimal Transport) formulates the post-processing of action segmentation as an optimal transport problem, specifically using a fused, unbalanced Gromov-Wasserstein formulation to decode temporally consistent segmentations from noisy affinity matrices \cite{xu2024temporally}. 
These advancements in temporal action segmentation, particularly the emphasis on multi-stage refinement and unsupervised optimal transport, provide valuable insights and potential techniques that can be adapted and applied to the problem of satellite PoL identification.


\section{Preliminaries}

\subsection{Problem Definition}\label{sec:sub-problem}

\noindent \textbf{Satellite Position:}
Satellite position can be represented as a multivariate time series, denoted as $\boldsymbol{x} = \left\{ p_t | t=1,\dots, T \right\}$, where $p_t$ is the position at a given time $t$, and $T$ is the length of the sequence.
Each position $p_t$ is a $d$-dimensional vector that records the satellite's location and movement in specific coordinate systems.
Typical coordinate systems include six Keplerian elements, geodetic coordinates (latitude, longitude, and altitude - LLA), and standard equatorial coordinate position and velocity vectors \cite{orbitalProjection}.

\noindent \textbf{Satellite PoL:}
Satellite pattern-of-life (PoL) refers to the behavior of a satellite over time, encompassing various operational activities and maneuvers. 
The station-keeping maneuvers are defined as the actions taken by a satellite to maintain its desired orbit, which is crucial for ensuring the satellite's operational efficiency and longevity.
The PoL of a satellite can be represented as a sequence of labels, denoted as $Y = \left\{ y_t | t=1,\dots,T \right\}$, where $y_t$ is the label at time $t$.
The PoL labels are categorized into two main types: station-keeping and Non-station-keeping behaviors.
Following the definition in previous studies \cite{challenge}, the station-keeping maneuvers are divided into three categories based on the propulsion system used: electric propulsion (EK), chemical propulsion (CK), and hybrid propulsion (HK).
Non-station-keeping behaviors are divided into initial drift (ID) and adjust drift (AD), where ID represents a satellite that leaves the desired orbit and starts drifting, and AD refers to any significant changes in longitude drift rate or direction for a drifting satellite.
Note that identification in EW and NS directions is independent, as EW and NS maneuvers correspond to different physical operations.

\noindent \textbf{Problem Statement (Satellite PoL Identification):}
The objective of satellite PoL identification is to determine the maneuvers of the satellite at each timestamp.
In other words, it is a sequence labeling problem where the goal is to predict the PoL label for each timestamp in the satellite position sequence.
Formally, the task is to learn a function $f: \boldsymbol{x} \rightarrow \hat{Y}$ that maps each position sequence $\boldsymbol{x}$ to a sequence of PoL labels $\hat{Y}$, where $\hat{Y} = \left\{ \hat{y}_t | t=1,\dots,T \right\}$ and $\hat{y}_t \in \mathcal{A}$, with $\mathcal{A} = \left\{ \textnormal{ID, AD, EK, CK, HK} \right\}$ representing the set of PoL classes.
The aim is to minimize the error between the prediction $\hat{Y}$ and the true PoL labels $Y$.
To ensure the high utility of the sequence labels, the prediction should not only have high point-wise accuracy at each timestamp but also perform well on the sequence level.

\subsection{Conditional Diffusion Probabilistic Model}\label{sec:sub-diffusion}

Diffusion probabilistic models have achieved state-of-the-art performance in various domains \cite{gong2022diffuseq, diffusionsuvery}, and their performance is further enhanced when conditioned on specific information \cite{ zhang2023adding, lin2023text, zhu2024controltraj}. 
Typical diffusion models comprise two processes: the forward (diffusion) process that gradually corrupts the original data into Gaussian noise, and the reverse (denoising) process that aims to recover the original data from the noise.

\noindent \textbf{Forward Process:} 
The forward process aims to transform the original data into a Gaussian noise distribution by sequentially adding noise.
Formally, this process can be represented as a Markov chain that transforms the original data $\mathbf{x}_0 \sim q(\mathbf{x}_0) $,
sampled from the data distribution $q(\mathbf{x}_0)$,
into a series of noisy data $\mathbf{x}_1, \mathbf{x}_2, \ldots, \mathbf{x}_S$, where $\mathbf{x}_S \sim \mathcal{N}(0, \mathbf{I})$.
At each diffusion step $ s $, noise is added according to a predefined variance schedule $\beta_s \in (0,1)$ that controls the amount of noise added at each step:
\begin{equation}
q(\mathbf{x}_s | \mathbf{x}_{s-1}) = \mathcal{N}(\mathbf{x}_t; \sqrt{1 - \beta_s} \mathbf{x}_{s-1}, \beta_s \mathbf{I}),   
\end{equation}
where $\mathcal{N}$ denotes a Normal distribution.
A key property of this process is that $\mathbf{x}_s$ at any arbitrary step $s$ can be sampled directly from $\mathbf{x}_0$.
For gradient-based optimization, the reparameterization trick is applied to efficiently sample from this distribution \cite{ddpm}.
By defining $\alpha_s = 1 - \beta_s$  and $ \bar{\alpha}_s = \prod _{i=1}^s (1 - \beta_s) $, the distribution of $x_s$ can be expressed as:
\begin{equation}
    \mathbf{x}_s = \sqrt{\bar{\alpha}_s} \mathbf{x}_0 + \epsilon \sqrt{1 - \bar{\alpha}_s},
\end{equation}
where $ \epsilon \sim \mathcal{N}(\mathbf{0}, \mathbf{I}) $.
Here, $\bar{\alpha}_s$ represents the signal rate (the amount of original signal remaining), while $1 - \bar{\alpha}_s$ is the noise rate. 
This equation shows that any noised sample $\mathbf{x}_s$ is simply a linear combination of the original data $\mathbf{x}_0$ and a random Gaussian noise vector $\epsilon$. 
As $s$ approaches $S$, $\bar{\alpha}_s$  approaches 0, and $\mathbf{x}_s$ becomes pure Gaussian noise.


\begin{figure*}
    \centering
    \includegraphics[width=\linewidth]{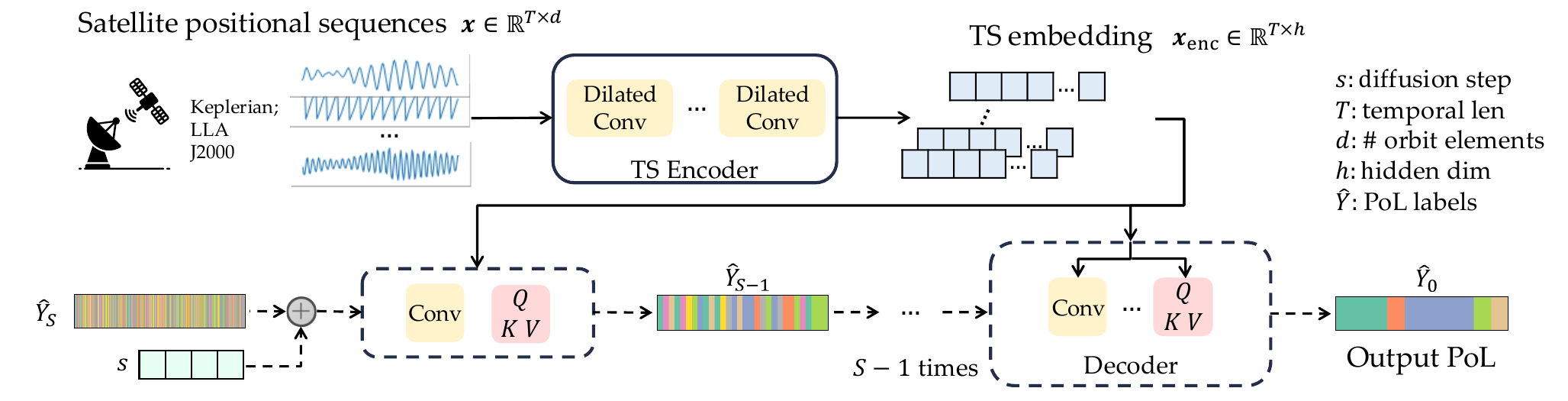}
    \caption{An overview of the proposed identification framework. 
    A time-series encoder first encodes satellite positional sequences into high-dimensional embeddings. 
    Then embeddings are fed into the decoder as conditional information in the denoising process via cross-attention.}
    \label{fig:overview}
\end{figure*}

\noindent \textbf{Reverse Process:}
The reverse process recovers the original data distribution from a Gaussian noise state. 
It starts from $x_S$ and progressively denoises the data using learned parameters. 
At each diffusion step $s$, the model predicts the mean and variance of the data distribution at the previous diffusion step $ s-1 $ conditioned on the current noisy data $ \mathbf{x}_s $ and some conditional information $ \mathbf{c} $. 
The conditional reverse process is described by the equation:
\begin{equation}
    p_\theta(\mathbf{x}_{s-1} | \mathbf{x}_s, \mathbf{c}) = \mathcal{N}(\mathbf{x}_{s-1}; \mu_\theta(\mathbf{x}_s, \mathbf{c}, s), \sigma_\theta^2(\mathbf{x}_s, \mathbf{c}, s) \mathbf{I}),
\end{equation}
where $ \mu_\theta $ and $\sigma_\theta$ are parameterized by a neural network with learnable parameters $\theta$. 
In our model, $\theta$ represents all the weights of the decoder network, detailed in \cref{sec:decoder}. 
The conditional information $\textbf{c}$ (the satellite's positional embedding) modulates this process by being integrated into the network, for example, through cross-attention layers, which allows the network's output to be guided by the input satellite data.


The core of the reverse process is learning to predict the noise $\epsilon$ or, equivalently, the original data $\mathbf{x}_0$ from the noisy input $\mathbf{x}_s$. 
It has been shown that the mean of the reverse step, $ \mu_\theta(\mathbf{x}_s, \mathbf{c}, s)$, can be derived by first training a model $ f_\theta(\mathbf{x}_s, \mathbf{c}, s)$ to predict $\mathbf{x}_0$. 
Once an estimate of $\mathbf{x}_0$ is obtained, the posterior distribution $q (\mathbf{x}_{s-1} | \mathbf{x}_s, \mathbf{x}_0)$ is tractable and can be used to sample the less-noisy data $\mathbf{x}_{s-1}$. 
The full sampling equation is \cite{ddim}:
\begin{equation}
\begin{aligned}
    \mathbf{x}_{s-1} =& \sqrt{\bar{\alpha}_{s-1}} f_\theta(\mathbf{x}_s, \mathbf{c}, s) + \\ 
    &\sqrt{1 - \bar{\alpha}_{s-1} - \sigma_\theta^2} \frac{\mathbf{x}_s - \sqrt{\bar{\alpha}_s}f_\theta(\mathbf{x}_s, \mathbf{c}, s)}{\sqrt{1-\bar{\alpha}_s}} + \sigma_s \epsilon.
\label{eq:reverse}
\end{aligned}
\end{equation}
This iterative process, starting from pure noise $\mathbf{x}_s$, gradually removes noise at each step, guided by the conditional information $\textbf{c}$, until a clean sample $\mathbf{x}_0$ is generated.

To sample from this distribution, we use the reparameterization trick, which allows the gradients to flow through the stochastic process. The noisy data at the previous diffusion step $ s-1 $ can be expressed as:
\begin{equation}
\mathbf{x}_{s-1} = \mu_\theta(\mathbf{x}_s, \mathbf{c}, s) + \sigma_\theta(\mathbf{x}_s, \mathbf{c}, s) \epsilon 
\end{equation}
where $ \mathbf{\epsilon} \sim \mathcal{N}(0, \mathbf{I}) $ is standard Gaussian noise. 
This formulation is what enables the generation of high-quality, conditioned samples. 
The model parameter $\theta$ is explicitly trained to reverse the noise addition process while being guided by the conditional input $\textbf{c}$. Therefore, the generated data is constrained to be consistent with both the learned data distribution (from reversing the diffusion) and the specific satellite positional sequence provided as a condition.

\section{Diffusion-based Method for PoL Identification}\label{sec:method}


In this section, we introduce the workflow of the proposed \SystemName, which identifies PoL in a point-wise sequence generation manner.
The overview of \SystemName is illustrated in \cref{fig:overview}.

Applying a diffusion model to satellite PoL identification is not a standard generation task and requires specific adaptations. 
Our method formulates the problem as a conditional sequence generation task, where the goal is to generate a sequence of PoL labels $Y$ conditioned on a satellite's positional data $x$. 
The key adaptations are in our conditioning scheme and the step-by-step refinement process for the discrete label sequence.

The generation process is guided by the satellite's orbital dynamics. 
As shown in \cref{fig:overview}, the multivariate positional sequence $x$ is first processed by a time-series encoder to produce a high-dimensional embedding $x_\textnormal{enc}$. 
This embedding serves as the conditional information $c$ for the diffusion decoder. 
It is integrated into the denoising network at each diffusion step via the cross-attention mechanism detailed in the decoder structure. 
This allows the model to continuously reference the relevant positional data when refining the PoL sequence.

Unlike diffusion models for images that operate on continuous pixel values, our model generates a sequence of discrete PoL labels. 
The forward process gradually adds noise to the ground truth label sequence $Y_0$. 
The reverse process, driven by the decoder, starts with random noise $Y_S$ and iteratively denoises it over $S$ steps to predict the final, well-structured PoL sequence. 
This iterative refinement, guided by the conditioning signal $x_\textnormal{enc}$, allows the model to produce temporally coherent and structured label sequences.
Details are elaborated in the following sections.

\subsection{Encoder Structure}

The encoder aims to learn hidden representations of the satellite sequences, which is crucial for accurate PoL identification.
The learned representations, which are utilized as conditional information in the denoising process, should encapsulate the temporal patterns and positional characteristics of the satellite sequences.
As the satellite positional sequences are multivariate and exhibit complex temporal patterns, we follow those well-developed modules in time-series representation learning \cite{yue2022ts2vec, zhang2024self} and design a multi-layered encoder structure.

\begin{figure}
    \centering
    \includegraphics[width=0.9\linewidth]{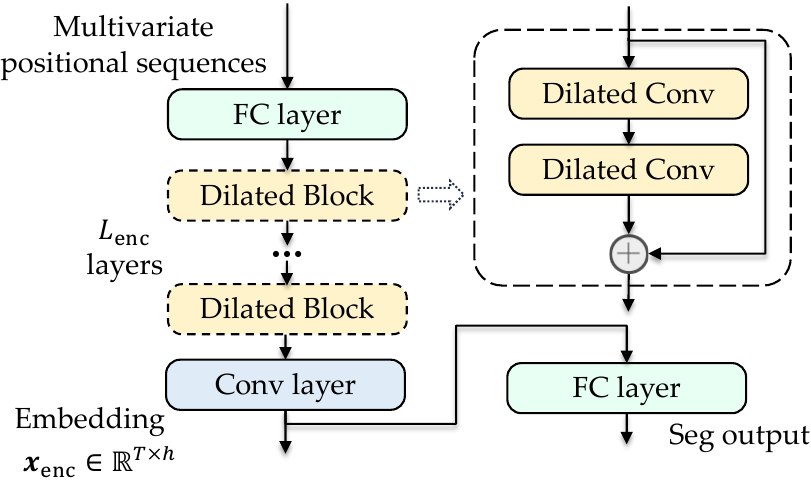}
    \caption{The proposed encoder structure. Dilated convolutional layers are applied to learn hierarchical temporal patterns.}
    \vspace{-10pt}
    \label{fig:encoder}
\end{figure}

The encoder structure of the proposed model, depicted in \cref{fig:encoder}, consists of a series of dilated convolutional blocks as the backbone.
The input of the encoder is the multivariate satellite positional sequences,  $\boldsymbol{x} \in \mathbb{R}^{T\times d}$, where $T$ is the temporal period length and $d$ is the number of orbit elements. 
The satellite positional sequence $\boldsymbol{x}$ is first passed through a fully connected (FC) layer to perform an initial transformation on the feature dimension.
Next, the transformed sequences are fed into $L_{enc}$ layers of dilated convolutional blocks. 
Each dilated block consists of two dilated convolutional layers, which are crucial for capturing long-range dependencies within the time-series data.
Mathematically, a dilated convolution with a learnable filter $\omega \in \mathbb{R}^K$ can be defined as:
\begin{equation}
    (\boldsymbol{x} *_{\delta } \omega)(t) = \sum_{i=0}^{K-1} \omega (i) \boldsymbol{x}(t - i \times \delta ),
    \label{eq:dilated_conv}
\end{equation}
where $\boldsymbol{x}$ is the input sequence, $K$ is the convolutional kernel size, and $\delta$ is the dilation factor.
With $\delta$, which controls the skipping distance of the convolution operation, the convolutional kernel picks inputs every $d$ step and applies the standard 1D convolution to the selected inputs.
The dilation factor increases exponentially with the depth of the dilated blocks, allowing the network to capture hierarchical temporal patterns without a significant increase in computational cost.
In this study, we set $\delta = 2^{l}$, where $l$ is the layer index of dilated blocks.
Additionally, a residual connection is applied in each block to facilitate the training process and prevent the vanishing gradient problem.

After passing through $L_{enc}$ layers of dilated blocks, the features are concatenated and processed by another convolutional layer to further refine the embeddings to the desired output dimension.
The output of this convolutional layer is a set of time-series embeddings, $\boldsymbol{x}_{\textnormal{enc}}^{T \times h}$, which encapsulate the temporal dynamics and positional information of the satellite and are fed into the decoder as conditional information in the denoising process.

Additionally, inspired by the design in the previous study \cite{diffact}, we include an extra output from the encoder. 
It is processed by a fully connected (FC) layer, which maps the embeddings to the class dimension. 
During training, this output is supervised using a cross-entropy loss, serving as an auxiliary loss to aid in model training, though it is not utilized during the inference process.


\subsection{Decoder Structure \label{sec:decoder}}

As introduced in \cref{sec:sub-problem}, the denoising process is achieved by a neural network module $f_\theta(\mathbf{x}_s, \mathbf{c}, s)$ that predicts label sequence $\boldsymbol{Y}_0$ at a given diffusion step $s$ with the noised state $\mathbf{x}_s$ and the conditional information $\mathbf{c}$.
The step-by-step denoising paradigm is similar to the sequence-level refinement in traditional segmentation methods \cite{mstcn++, ding2024temporal}. 
However, while those methods rely on different network designs or manually crafted rules, the denoising process can achieve this refinement in an end-to-end manner.

\begin{figure}
    \centering
    \includegraphics[width=0.8\linewidth]{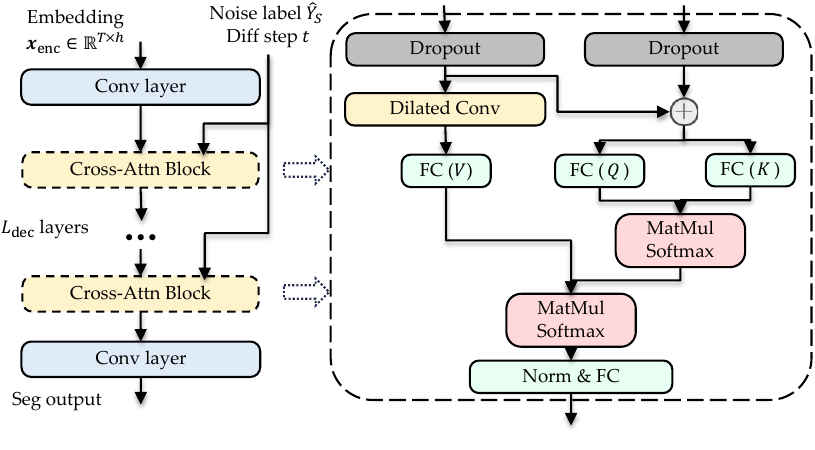}
    \caption{The proposed decoder structure. Cross-attention block is designed to integrate embedding and noised state.}
    \vspace{-10pt}
    \label{fig:decoder}
\end{figure}

In this study, we customize the decoder structure and identify PoL labels in a point-wise sequence generation way.
Specifically, each diffusion state is a noised PoL label sequence $\hat{Y}_s$, and the embedding of the positional sequence $\boldsymbol{x}_{\textnormal{enc}}$ is utilized as conditional information $\mathbf{c}$.
The decoder structure of the proposed model, illustrated in \cref{fig:decoder}, utilizes a combination of convolutional and cross-attention blocks.
Inputs of the decoder are grouped into twofold: the noised PoL label sequence ${Y}_s$ combined with the diffusion step $s$, and the conditional information $\boldsymbol{x}_{\textnormal{enc}}$.
The diffusion step $s$ is first embedded into the same dimension as the noised state $\hat{Y}_s$ with sinusoidal embedding \cite{vaswani2017attention}, and then concatenated with the noised state, denoted as $h_s$.
The embedding from the encoder first passes through a convolutional layer, denoted as $h_c$.

Subsequently, the decoder incorporates $L_{\textnormal{dec}}$ layers of cross-attention blocks. 
Each cross-attention block includes a dilated convolution layer, which captures hierarchical dependencies within the embedding from the encoder. 
Additionally, cross-attention mechanisms are applied to integrate the conditional information from the encoder and the noised state.
\begin{equation}
    \begin{aligned}
        \boldsymbol{Q} &= \boldsymbol{W}_q \cdot \left(h_c \oplus h_s \right), \\
        \boldsymbol{K} &= \boldsymbol{W}_k \cdot \left(h_c \oplus h_s \right), \\
        \boldsymbol{V} &= \boldsymbol{W}_v \cdot \textnormal{Conv} \left(h_c\right),
    \end{aligned}
\end{equation}
where $h_c$ is the conditional embedding, $h_s$ is the state embedding, and $\boldsymbol{W}_q, \boldsymbol{W}_k, \boldsymbol{W}_v$ are learnable weight matrices.
Then, the output of cross-attention is calculated as: 
$h = \text{softmax}(\frac{\boldsymbol{Q} \cdot \boldsymbol{K}^T}{\sqrt{d_k}}) \cdot \boldsymbol{V}$. 
To prevent overfitting and improve generalization, dropout layers are applied before and after the dilated convolution operations.

Finally, the refined embeddings from the cross-attention blocks are passed through a final convolutional layer, which maps the learned features to PoL classes.
And a softmax layer is applied to calculate the probability at each time step over different PoL categories, $p_i = \textnormal{softmax}\frac{e^{z_i}}{\sum_{C} e^z_i}$.
This output, i.e., the denoised sequence, $p \in \mathbb{R}^{T \times C}$ represents the PoL labels $\hat{Y}_s$, indicating the identified behaviors of the satellite over time. 

A key aspect of our approach is the application of a continuous Gaussian diffusion process to the PoL label sequences, which are inherently discrete. 
We represent the label sequence $Y$ as discrete points in a continuous high-dimensional space. 
This modeling choice, while seemingly counter-intuitive, is an established and empirically effective technique for applying diffusion models to discrete data domains, such as temporal action segmentation \cite{ding2024temporal}. 
During the reverse process, the decoder learns to map the noisy continuous vectors back towards the vertices of the probability simplex, which are then interpreted as valid categorical probabilities via the final softmax layer.

\begin{algorithm}[t]
    \caption{Train and Inference Processes of \SystemName}\label{alg:train}
    \raggedright
  
  \textbf{Training Process:}
  \begin{algorithmic} [1]
      \For{ $i = 1, 2, \ldots,$}
        \State Positional sequence $\boldsymbol{x}$, PoL label $Y$
        \State Get embedding $\boldsymbol{x}_{\textnormal{enc}} $ from the encoder $\theta_{\textnormal{enc}}$
        \State Get masked embedding $\boldsymbol{x}_{\textnormal{enc}} = \boldsymbol{x}_{\textnormal{enc}} \odot M$
        \State Sample ${Y}_0$ from label distribution $q(Y)$,
        \State Sample diffusion step $s \sim \operatorname{Uniform}(\left\{1, \ldots, S \right\})$, $\epsilon \sim \mathcal{N}(0, \mathbf{I})$
        \State $Y_s = \sqrt{\hat{\alpha}_s} Y_0 + \epsilon \sqrt{1-\bar{\alpha}_s} $
        \State Denoise with the decoder $\theta_{\textnormal{dec}}$, $\hat{Y}_0 =f_{\theta_{\textnormal{dec}}} (Y_s, \boldsymbol{x}_{\textnormal{enc}},s)$
        \State Update gradient $\nabla \theta_{\textnormal{enc}}$, $\nabla \theta_{\textnormal{dec}}$, with loss $\mathcal{L} = \mathcal{L}_{ce} + \mathcal{L}_{smo} + \mathcal{L}_{bd}$
      \EndFor
  \end{algorithmic}
  
  \textbf{Inference Process:}
  \setcounter{algorithm}{9}
  \begin{algorithmic}[1]
      \makeatletter
      \setcounter{ALG@line}{10}
      \State Positional sequence $\boldsymbol{x}$
      \State Get embedding $\boldsymbol{x}_{\textnormal{enc}} $ from the encoder $\theta_{\textnormal{enc}}$
      \State Sample $\hat{Y}_S \sim \mathcal{N}(0,\mathbf{I})$
      \For{ $s = S, S - \Delta s, \ldots, 1$}
        \State Compute $\hat{Y}_s$ according to \cref{eq:reverse}
      \EndFor \\
      \Return $\hat{Y}_0$
  \end{algorithmic}
\end{algorithm}
\vspace{-10pt}

\subsection{Training and Inference}\label{sec:sub-training}

With the introduced encoder and decoder structures, the diffusion model generates PoL label sequence based on the noised state and the conditional information.
The training and inference processes of the proposed model are detailed in \cref{alg:train}.

\textbf{Training process}:
The training process begins by embedding the positional sequence $x$ into $x_{\textnormal{enc}}$ using the encoder $\theta_{\textnormal{enc}}$.
Then, we apply the condition masking strategy to control the conditional information, which functions as a form of classifier-free guidance \cite{ho2022classifier}. 
This strategy involves randomly selecting one of several mask types to apply to the conditional embedding $x_{\textnormal{enc}}$ during each training step \cite{diffact}.
Each mask $M$ is designed to teach the model a different structural prior about the data.
i) No Masking: This is the standard conditional case where the full embedding $x_{\textnormal{enc}}$ is passed to the decoder. 
It allows the model to learn the direct mapping from orbital data to the PoL sequence.
ii) Position Prior Mask: This is an all-zero mask that completely blocks the conditional embedding. It forces the model to learn the unconditional distribution of PoL sequences (e.g., typical maneuver durations and frequencies) without relying on any specific input data.
iii) Boundary Prior Mask: This mask removes the features specifically at the transition points between different PoL states. This encourages the model to use the more stable context from within a segment to accurately identify ambiguous state boundaries.

Next, for a diffusion step $s$, a noised PoL label sequence $y_s$ is sampled. 
The decoder denoises $\hat{y}_s$ to predict $\hat{y}_0$ by leveraging the conditional information from $x_{\text{enc}}$. 
The encoder and decoder are optimized jointly with a combined loss function $L$, which are detailed as follows.

\textit{Cross-Entropy Loss}:
Conventional cross-entropy loss is applied as point-wise supervision, which minimizes the negative log-likelihood of the ground truth PoL class for each timestep. 
The cross-entropy loss $\mathcal{L}_{ce}$ is defined as:
\begin{equation}
    \mathcal{L}_{ce} = - \frac{1}{TC} \sum_{T} \sum_{C} Y_{t,c} \log p_{t,c}    
\end{equation}
where $T$ is the temporal length, $C$ is the number of PoL categories, $Y$ is the ground truth label, and $p_{t,c}$ is the predicted probability for class $c$ at time step $t$.

\textit{Smooth Loss}:
A smooth loss is applied to maintain consistency in predictions over consecutive frames.
The loss $\mathcal{L}_{smo}$ is calculated by minimizing the difference in the log-likelihoods between adjacent frames \cite{mstcn, mstcn++}:
\begin{equation}
    \mathcal{L}_{smo} = \frac{1}{(T-1)C} \sum_{T-1} \sum_{C} \left( \log p_{t,c} - \log p_{t+1,c} \right)^2    
\end{equation}

\textit{Boundary Alignment Loss}:
Identifying the transition boundaries between different PoL classes is crucial for PoL identification, as well as other sequence segmentation tasks.
The boundary probabilities are computed by taking the dot product of the action probabilities from neighboring frames in $p_s$, and the alignment is achieved using a binary cross-entropy loss. 
The boundary alignment loss $\mathcal{L}_{bd}$ is defined as \cite{diffact}:
\begin{equation}
\begin{aligned}
    \mathcal{L}_{bd} = \frac{1}{T-1} \sum_{T-1} \Big[ -\bar{B}_t \log (1 - p_{t} \cdot p_{t+1}) \\
    - (1 - \bar{B}_t) \log (p_{t} \cdot p_{t+1}) \Big]    
\end{aligned}
\end{equation}
where $\bar{B} \in \mathbb{B}^{T}$ is the smoothed boundary sequence derived from the ground truth.



\noindent  \textbf{Inference process}:
As described in \cref{alg:train}, the inference process is guided by the learned representation $x_{\text{enc}}$ from the encoder.
The initial noisy PoL label sequence $\hat{y}_S$ is generated by sampling from a Gaussian distribution $\mathcal{N}(0, I)$.
Subsequently, the model iterates through each diffusion step, starting from $S$ down to 1. 
At each step $s$, the decoder utilizes the current noisy PoL label $\hat{y}_s$, the conditional embedding $x_{\text{enc}}$, and the diffusion step information to refine the sequence according to \cref{eq:reverse}. 
The denoising process progressively reduces the noise, moving closer to the ground truth.

\section{Experiments}\label{sec:exper}

In this section, we first describe the datasets, baseline methods, and
evaluation metrics used in our experiments. 
We then comprehensively evaluate \SystemName on real-world data
to answer the following research questions:

\begin{itemize}
    \item \textbf{RQ1:} How does \SystemName perform in identifying satellite pattern-of-life compared to state-of-the-art baseline methods?
    
    \item \textbf{RQ2:} How robust is \SystemName when dealing with satellite positional data of varying quality and sampling rates from different ephemerides sources?

    \item \textbf{RQ3:} How do different components and design choices of \SystemName contribute to its overall performance?
    
    \item \textbf{RQ4:} What are the computational costs of \SystemName compared to baseline methods, and is it practical for real-world deployment?

    \item \textbf{RQ5:} How sensitive is \SystemName to different hyperparameters, such as noise level, diffusion steps, and sequence length?

\end{itemize}

\subsection{Settings}

\subsubsection{Dataset}

We evaluate \SystemName on the Satellite Pattern-of-Life Identification Dataset (SPLID), designed by MIT ARCLab to support satellite behavior characterization studies \footnote{\url{https://eval.ai/web/challenges/challenge-page/2164/overview}}.
The dataset contains both synthetic data generated by a high-fidelity satellite simulation tool and real-world data derived from satellite tracking records \cite{challenge}. 
The position is recorded every two hours with 15 orbital parameters, including orbital elements, geodetic positions and J2000 position and velocity.
PoL labels in EW and NS directions are provided for each satellite, details are summarized in \cref{tab:dataset}.

\begin{table}
    \small
    \centering
    \caption{SPLID dataset statistics}
    \label{tab:dataset}
    \begin{tabular}{ccccc}
    \toprule
    Direction & \# Object & \# Period & \# PoL label & \# Transition / Object \textsuperscript{a} \\
    \cmidrule(lr){1-5}
    EW & \multirow{2}{*}{1900} & \multirow{2}{*}{6 months} & 5 & 1.69 \\
    NS & & & 4 & 1.13 \\
    \bottomrule
    \end{tabular}
    \vspace{3pt}

    \begin{minipage}{0.9\linewidth} 
        \small 
        \textsuperscript{a} '\# Transition/Object' refers to the average number of PoL state changes per satellite object in the dataset.
    \end{minipage}

\end{table}

\begin{figure}
    \centering
    \subfloat[PDF and CDF of the low sampling interval. The average interval is 13.7 hrs.]{
        \includegraphics[width=0.48\linewidth]{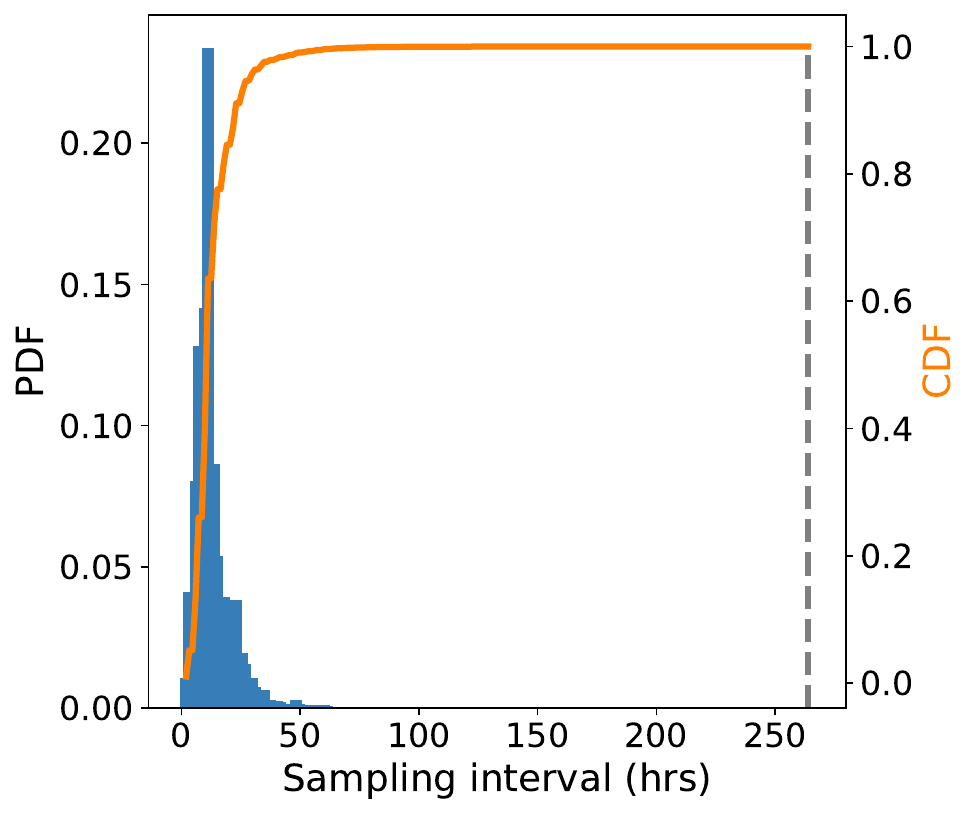}
        \label{fig:sampling-pdf}
    }
    \hfill
    \subfloat[Example for positional sequences (Eccentricity) with low sampling rate.]{
        \includegraphics[width=0.4\linewidth]{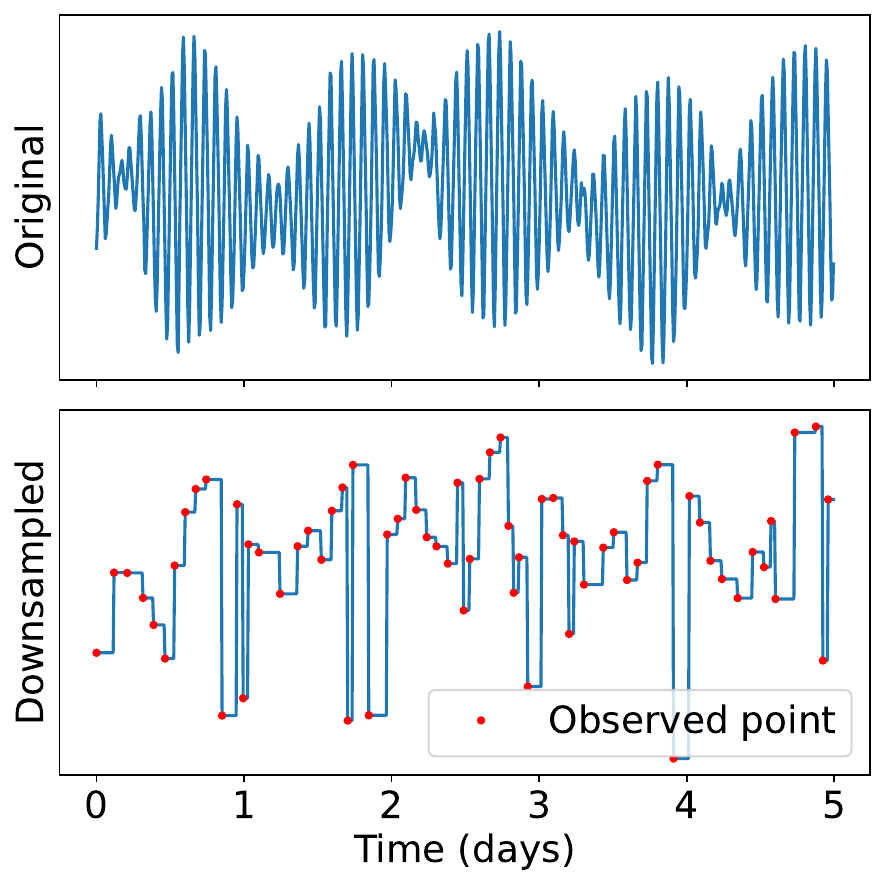}
        \label{fig:sampling-eg}
    }
    \caption{Positional sequences with low sampling rate.}
    \label{fig:sampling}

\end{figure}

In addition to the original high-fidelity dataset, we consider a more realistic scenario where satellite positional data varies in quality and resolution based on the source of ephemerides. 
When using publicly available TLE-derived data rather than precision simulation tools and VCM-derived ephemerides, the inherently lower sampling rate and accuracy presents significant challenges for PoL algorithm deployment \cite{challengereport}. 
This reduced data quality can result in incomplete or imprecise positional sequences that fail to capture subtle maneuvers, thereby substantially affecting the identification performance of traditional methods that were developed using high-precision data sources. 
In \cref{fig:sampling-pdf}, we present the Probability Density Function (PDF) and Cumulative Distribution Function (CDF) of satellite position data sampling intervals of TLE-derived data in the dataset. 
The sampling interval distribution shows that most intervals are concentrated in a short time range ($\sim$ 12 hours), but there are also instances of significantly longer intervals (more than 7 days), indicating substantial variability in sampling rates. 
\cref{fig:sampling-eg} illustrates an example of a positional sequence with both original and low sampling rates. 
The red dots represent the observed positions, and the sequence is imputed to its original sampling rate using forward padding, which does not incorporate any external knowledge or assumptions.
It is evident that the reduced sampling rate severely truncates the original positional sequence, resulting in a markedly different time-series pattern. 
Thus, we evaluate the PoL identification performance with both original and low sampling rate positional sequences.

\subsubsection{Evaluation metrics}

To fully evaluate the quality of PoL identification, we follow the practice in sequence segmentation tasks \cite{ding2024temporal, diffact}, and use three evaluation metrics at different levels.
Specifically, let $Y$ be the ground truth label sequence and $\hat{Y}$ be the output of the model, the metrics are defined as follows:

\textbf{Accuracy (Acc)}:
Accuracy is a point-wise metric that measures the proportion of correctly identified PoL labels out of the total labels.
    \begin{equation}
        \textnormal{Acc} = \frac{\sum_{i=1}^{n} \mathbb{I}(Y_i = \hat{Y}_i)}{n} \times 100\%,
    \end{equation}
where $\mathbb{I}(\cdot)$ is the indicator function that equals 1 when the condition is true and 0 otherwise, and $n$ is the total number of labels.
    
\textbf{Edit score (Edit)}:
The edit score is a sequence-level metric that quantifies the similarity between two sequences based on the Levenshtein distance, which is the minimum number of actions required to transform one sequence into another.
The accumulated distance value $e_{i,j}$, representing the Levenshtein distance between the first $i$ labels of the ground truth sequence $Y$ and the first $j$ labels of the predicted sequence $\hat{Y}$, is defined recursively as:
\begin{equation}
    e_{i,j} = \left\{ 
        \begin{array}{ll}
        \max(i, j), & \textnormal{if } \min(i, j) = 0 \\
        \begin{aligned}
            \min \big( & e_{i-1,j} + 1, e_{i,j-1} + 1, \\
                       & e_{i-1,j-1} + \mathbb{I}(Y_i \neq \hat{Y}_j) \big) 
            \end{aligned} & \textnormal{otherwise}

        \end{array}
        \right.
\end{equation}
Here, the $\min$ function considers the cost of three possible operations: deleting a label from $Y (e_{i-1,j} + 1)$, inserting a label into $Y (e_{i,j-1} + 1)$, or substituting a label $(e_{i-1,j-1} + I(...))$.
Then, the final edit score is calculated based on the total edit distance for the full sequences, $e = e[|Y|, |\hat{Y}|]$, normalized by the maximum sequence length:
\begin{equation}
    \textnormal{Edit} = \left( 1 - \frac{e[|Y|, |\hat{Y}|]}{\max(|Y|, |\hat{Y}|)} \right) \times 100\%.
\end{equation}

\textbf{F1-score (F1)}:
F1-score is a segment-level metric that compares the Intersection over Union (IoU) of each segment with respect to the corresponding ground truth based on a threshold $\tau$.
A segment is considered a true positive if its IoU with respect to the ground truth exceeds the threshold. 
If there is more than one correct segment within the span of a single ground truth action, only one segment is considered a true positive, and the others are marked as false positives. 
Based on the true and false positives, as well as false negatives (missed segments), the precision and recall are computed.
The following formulas are shown for a single class $c$ in a one-vs-rest manner, where the indicator function checks for membership in that class:
\begin{equation}
\begin{aligned}
    \textnormal{Precision} &= \frac{\sum_{i=1}^{n} \mathbb{I}(Y_i = \hat{Y}_i \textnormal{ and } \hat{Y}_i = c)}{\sum_{i=1}^{n} \mathbb{I}(\hat{Y}_i = c)} \\
   \textnormal{Recall} &= \frac{\sum_{i=1}^{n} \mathbb{I}(Y_i = \hat{Y}_i \textnormal{ and } Y_i = c)}{\sum_{i=1}^{n} \mathbb{I}(Y_i = c)}
\end{aligned}
\end{equation}
and the F1-score is computed as the harmonic mean of precision and recall.
Follow common practices \cite{ding2024temporal,diffact}, we set the threshold $\tau$ to ${0.1, 0.25, 0.5}$.

\subsubsection{Implementation details}
In this study, positional records of 1900 satellites are applied.
We select six Keplerian elements and geodetic positions as input features, which are 9-dimension time-series sequences.
For each satellite, we use the full six-month observation sequence as input, resulting in a sequence length of $T=2172$. 
This approach is crucial as PoL identification relies not only on local temporal features but also on the contextual information from adjacent PoL states throughout the sequence.
During the training stage, the dataset is randomly divided into training and testing sets with a ratio of 80\% and 20\%, respectively.
As maneuvers in EW and NS directions correspond to motion states, we treat PoL identification in the EW and NS directions as two separate tasks.

\begin{table*}
    \small
    \centering
    \caption{Performance comparison of \SystemName and baseline methods in EW and NS directions.}

    \begin{tabular}{ccccccc} 
    \toprule
    \multirow{2}{*}[\multirowoffset]{Method}  & \multicolumn{3}{c}{EW direction} & \multicolumn{3}{c}{NS direction} \\ 
    \cmidrule(lr){2-4}\cmidrule(lr){5-7}
    & F1 (\%) @ \{10,25,50\}  ($\uparrow$) & Edit (\%)  ($\uparrow$) & Acc (\%) ($\uparrow$) & F1 (\%)  @\{10,25,50\} ($\uparrow$) & Edit (\%) ($\uparrow$) & Acc (\%) ($\uparrow$) \\ 
    
    \cmidrule(lr){1-7}
     Heuristic      & 24.46  /  22.34  /  21.86  &  59.19 &  38.83  & 43.18 / 39.17 / 32.49     &  37.06  &  38.41  \\
     ClaSP    &  28.66 / 20.47 / 11.69   & 24.91
     &  41.66     &  36.66 / 26.72 / 14.07     & 37.07     &  42.23     \\
     RF    & 42.02 / 41.71 / 40.88  &  74.82 &  \textbf{95.50}  & 34.30 / 33.96 / 33.11  &  70.55 &  95.60  \\
     CNN & 39.39 / 34.02 / 27.51 & 41.13 & 76.43 & 58.50 / 57.54 / 54.52 & 66.40 & 91.55 \\
    TS2Vec & 19.61 / 18.98 / 17.80 & 68.73 & 88.61 &  23.61 / 23.18 / 22.40 & 56.52 & 95.03 \\
    MS-TCN++ & 57.42 / 52.40 / 44.92& 60.54 & 78.13 & 36.67 / 33.44 / 29.07 & 39.29 & 84.27 \\
    ASFormer & 15.07 / 14.28 / 13.57 & 45.80  & 80.81 & 19.87 / 19.35 / 18.38 & 40.07 & 87.34 \\

     \cmidrule(lr){1-7} 
     Expert-ML    & \textbf{93.92} / \textbf{93.71} / \textbf{93.40}  &  \textbf{93.79} &  92.94  & \textbf{95.69} / \textbf{95.19} / 93.69  & 93.97  &  95.21  \\
    \SystemName & 91.68 / 91.00 / 89.92 & 93.00 & 94.10 & 94.79 / 94.67 / \textbf{94.08} & \textbf{96.72} & \textbf{98.20} \\
    \bottomrule
    \end{tabular}
    \label{tab:comp}

    \vspace{2pt}
    Bold indicates the best performance over the baselines. $\uparrow$: higher is better.

\end{table*}

For the model hyperparameters, 9 layers of dilated convolutional blocks are concatenated in the encoder, with a hidden dimension of 64 and an initial dilation rate of 2.
The convolutional kernel size is set to 12 to capture the daily periodicity of the satellite. 
For the decoder, 3 layers of cross-attention and convolutional blocks are stacked, with a hidden dimension of 64. 
The dropout rate is set to 0.1 for both the encoder and decoder. 
The smooth loss is applied to minimize the difference in the log-likelihoods between adjacent frames \cite{mstcn++, diffact}.
The encoder and decoder are trained in an end-to-end manner using the Adam optimizer with a batch size of 4.
The learning rate is initialized at 5e-4 and follows a decayed rate at 1e-5.
The number of diffusion steps is set to 1000, with 25 steps utilized at inference based on a sampling strategy with skipped steps. 
All experiments are implemented in PyTorch and conducted on a server with NVIDIA A30 GPU. 
The implementation is available for reproducibility 
\footnote{\url{https://github.com/yeyongchao/satellite-PoL-diffusion}}.

\subsubsection{Baselines}
We compare the proposed \SystemName with the following baselines.

\begin{itemize}
    \item \textbf{Heuristic:} \cite{challenge} A heuristic method is developed along with the satellite PoL dataset, which is a combination of rules based on satellite maneuvers.

    \item \textbf{ClaSP:} \cite{clasp} ClaSP is an unsupervised change point detection method for invariant time-series data.
    We select longitude and inclination and apply ClaSP for EW and NS detection, respectively.
    
    \item \textbf{Random Forest (RF):} We apply a 7-day sliding window on the positional sequences and train an RF classifier to predict PoL labels at each timestamp.

    \item \textbf{CNN:} We train a three-layer convolutional neural network (CNN) that takes the positional sequences as input and generates a sequence of PoL labels.
    
    \item \textbf{TS2Vec:} \cite{yue2022ts2vec} TS2Vec is a universal time-series representation learning model. We apply its encoder module to learn the representation of the positional sequences.
    
    \item \textbf{MS-TCN++}:  \cite{mstcn++} MS-TCN++ is a multi-stage temporal convolutional network that generates an initial prediction and then iteratively refines it using subsequent stages with dual dilated convolutions to capture both local and global context.
    
    \item \textbf{ASFormer}: \cite{yi2021asformer} ASFormer is a Transformer-based model for action segmentation that incorporates local connectivity inductive biases and a hierarchical attention mechanism for efficient processing and iterative refinement.
    

    \item \textbf{Expert-ML:} We develop a strong, hybrid baseline that mirrors a sophisticated expert-driven approach, ranking 6 out of over 100 participant teams in the challenge \cite{challengereport}.
    i) For EW direction, drift events are first identified using a rule-based method that detects sharp ``step changes'' in eccentricity. 
    Following this, a Random Forest classifier is trained on 1-day sliding windows of orbital elements to identify the specific type of station-keeping.
    ii) For NS direction, a similar hybrid approach is used, but with different features and rules. 
    NS drift is identified based on rules related to the expected periodicity of maneuvers. 
    The subsequent station-keeping classifier uses a 14-day sliding window on equinoctial coordinates, which are better suited for capturing long-term NS dynamics.
    iii) Finally, a cross-verification step refines the outputs from both pipelines to ensure consistency and handle exceptional cases. This workflow represents a powerful but complex system that relies heavily on domain-specific feature engineering and rules.

\end{itemize}

\subsection{Overall accuracy (RQ1)}\label{sec:exp-acc}

\subsubsection{Quantitatiev analysis}
\cref{tab:comp} presents a comprehensive performance comparison of various methods in both the EW and NS directions using the F1 score at different thresholds (10, 25, 50), edit score, and accuracy.
Generally, all methods perform better in the NS direction than in the EW direction, because the EW direction contains five types of PoL, while the NS direction only contains four.
The proposed method, \SystemName, and Expert-ML achieved similar performance, indicating that the proposed method can achieve performance comparable to complex rules in an end-to-end manner, eliminating the dependence on domain expert knowledge.
The Expert-ML high F1 score in the EW direction is attributable to its hand-crafted rules, like 'step change' detection, which excel on clean, high-resolution data. However, this reliance on high-fidelity features is a critical vulnerability. As shown in \cref{sec:exp-gen}, these rules fail when features are obscured by lower sampling rates, causing the system's performance to collapse.
\SystemName shows superior performance across all metrics compared to other baselines. 
Especially for segment and sequence-level metrics, \SystemName achieves steady and accurate performance with F1 and edit scores above 90\% in both directions.
State-of-the-art temporal segmentation methods, such as MS-TCN++ and ASFormer, do not achieve satisfactory performance when directly applied to this task. Although MS-TCN++ shows a high F1 score in the EW direction, its overall accuracy is limited. This suggests that these general-purpose models struggle with the unique challenges of satellite data, such as strong temporal periodicity, and would require significant customization to be effective.
Although some baseline methods, such as TS2Vec and RF, achieved accurate point-wise accuracy, their performance at the segment and sequence level was poor, with F1 and edit scores ranging from 30\% to 70\%.
Heuristic and CLaSP performed poorly in all metrics, indicating that those universal methods are not directly capable of PoL identification.
These significant scores highlight the effectiveness of \SystemName in handling complex satellite positional data and refining the identification process, thereby achieving accurate and reliable PoL identification.

\begin{figure}
    \centering
    \includegraphics[width=\linewidth]{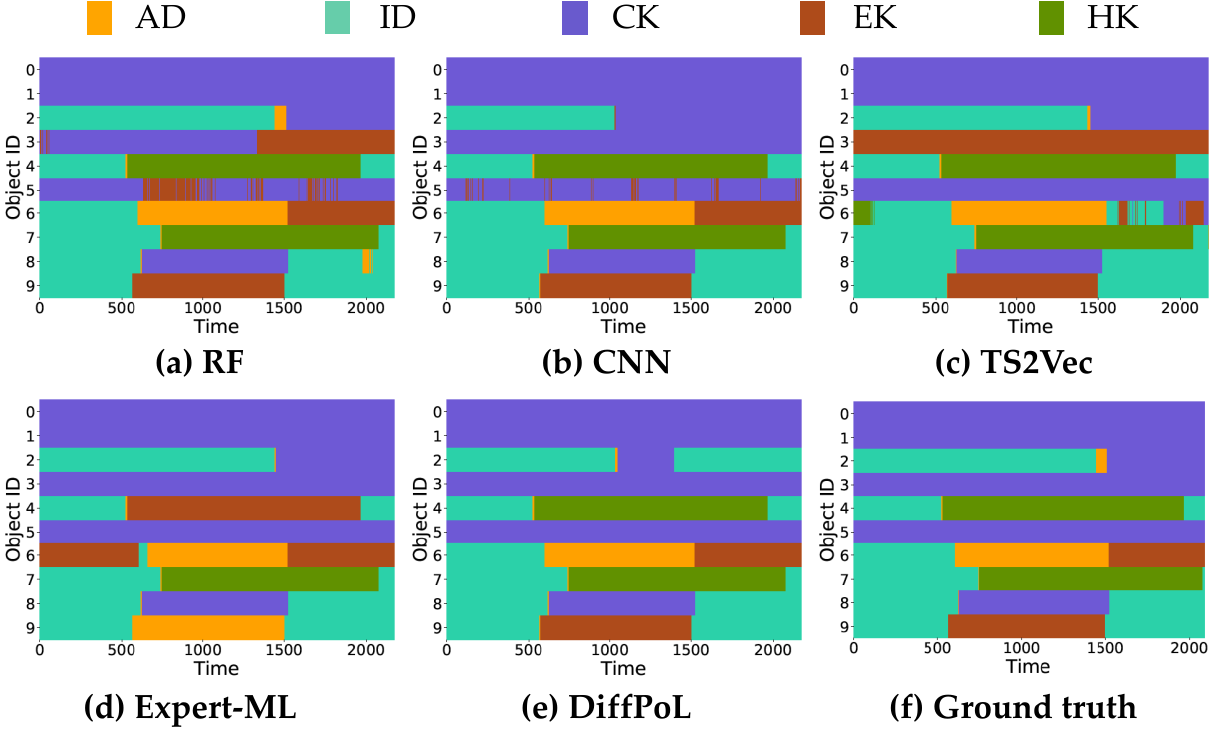}
    \caption{Visualization of identification results in EW direction.
    (a-c) Identifications are fragmented at transition boundaries. (d,e) Identifications are clear and structured, but (d) is less accurate.
    }
    \label{fig:heatmap}
\end{figure}

\begin{figure*}
    \centering
    \includegraphics[width=0.9\textwidth]{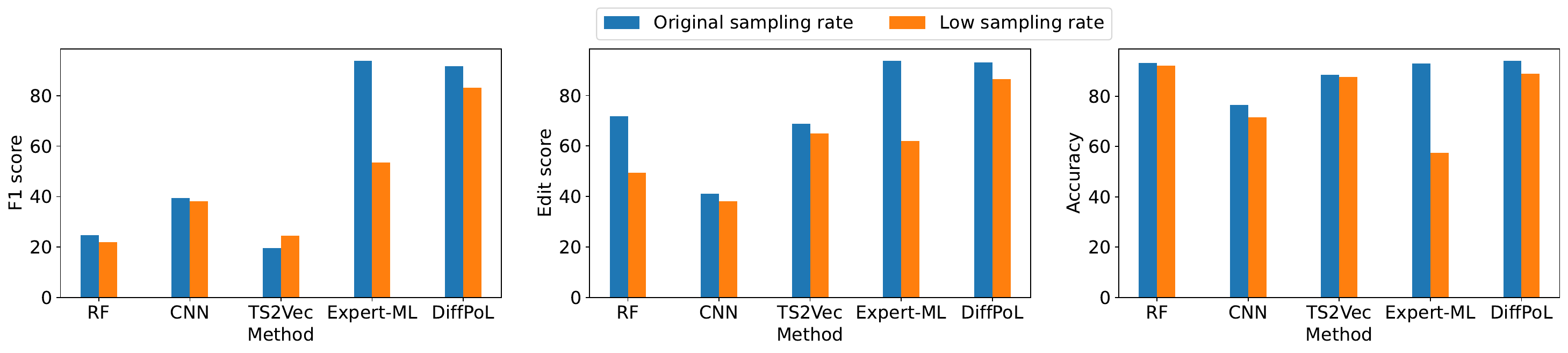}
    \caption{Identification performance comparison with different sampling rates in EW direction.}
    \label{fig:sampling-res-EW}
\end{figure*}

\begin{figure}
    \centering
    \includegraphics[width=0.8\linewidth]{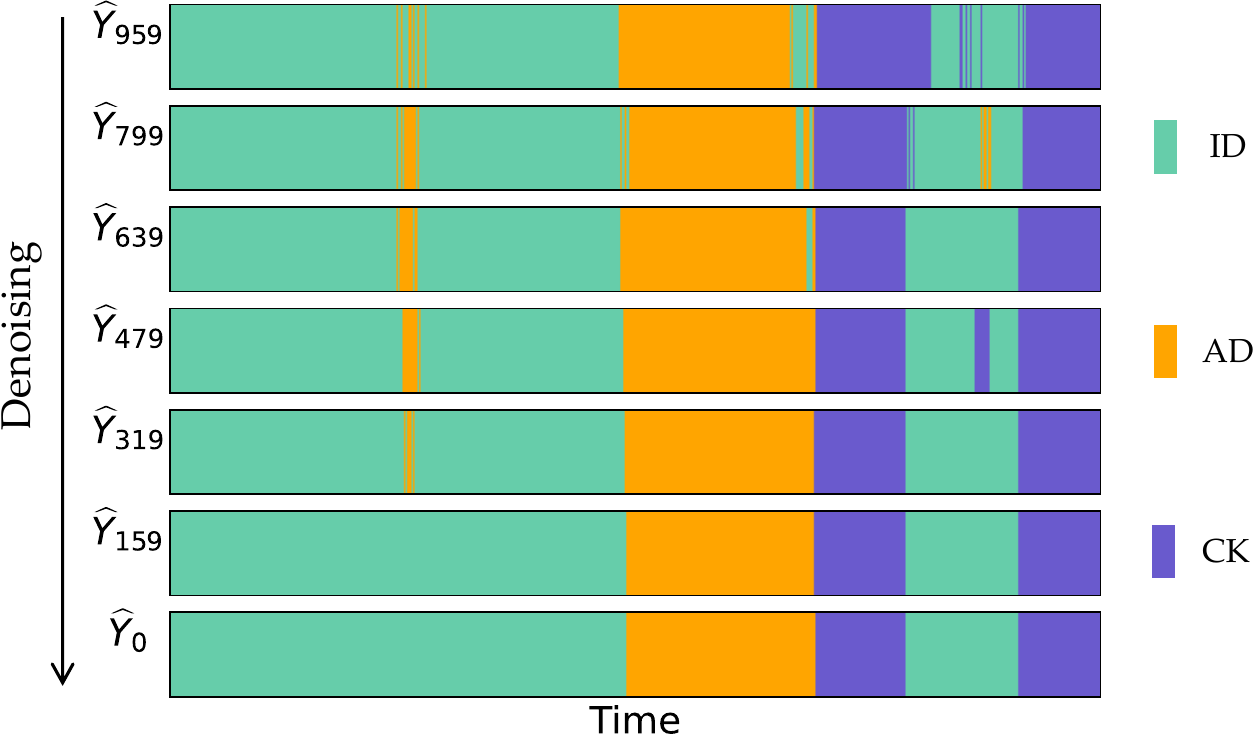}
    \caption{Visualization of the denoising process.}
    \label{fig:diffsteps}
\end{figure}

\subsubsection{Visualization analysis}
\cref{fig:heatmap} visualizes the identification results of 10 randomly selected objects in EW direction compared to the ground truth.
The identification results of RF, CNN and TS2Vec, \cref{fig:heatmap} (a-c), although achieving high point-wise accuracy, are fragmented, particularly at the boundaries of different states, leading to unnecessary segmentation.
Such fragmentation often requires manual refinement, which is labor-intensive and time-consuming.
\SystemName, as well as Expert-ML, shows clear and structured identification sequences, as shown in \cref{fig:heatmap} (d,e).
These structured identifications are practicable for further analysis and can be directly used for downstream applications.
Despite this structural clarity, the figure also reveals Expert-ML's critical weakness: brittleness. 
On objects 4, 6, and 9, the system fails entirely because its hand-crafted rules are tuned for conventional maneuvers and cannot adapt to satellites with more subtle or atypical control strategies.
The visualization results, consistent with the quantitative analysis of F1 and edit scores, highlight the superiority of \SystemName in identifying satellite PoL in an end-to-end manner without the need for further refinement of the sequence identification results.

We further select an object and visualize its denoising process at different steps in \cref{fig:diffsteps}, where different color represent different PoL labels.
From $\hat{Y}_{959}$ to $\hat{Y}_0$, is the beginning to end of the denoising process.
It can be observed that, at the beginning of the denoising process, $\hat{Y}_{959}$, the sequence is heavily noisy and fragmented with many misclassifications.
As the denoising process progresses, the identification gradually becomes clearer and more structured, particularly the fluctuations at the boundaries of different states are gradually refined. 
This is consistent with the quantitative results, where such a refinement process significantly reduces unnecessary segmentation, improving identification in sequence-level and showing a significant improvement in F1 and edit scores. 
This visualization demonstrates the effectiveness of the denoising process in refining the sequence to achieve accurate PoL identification, which aligns well with the motivation of applying the diffusion process in \SystemName.

\begin{figure}
    \centering    
    \includegraphics[width=\linewidth]{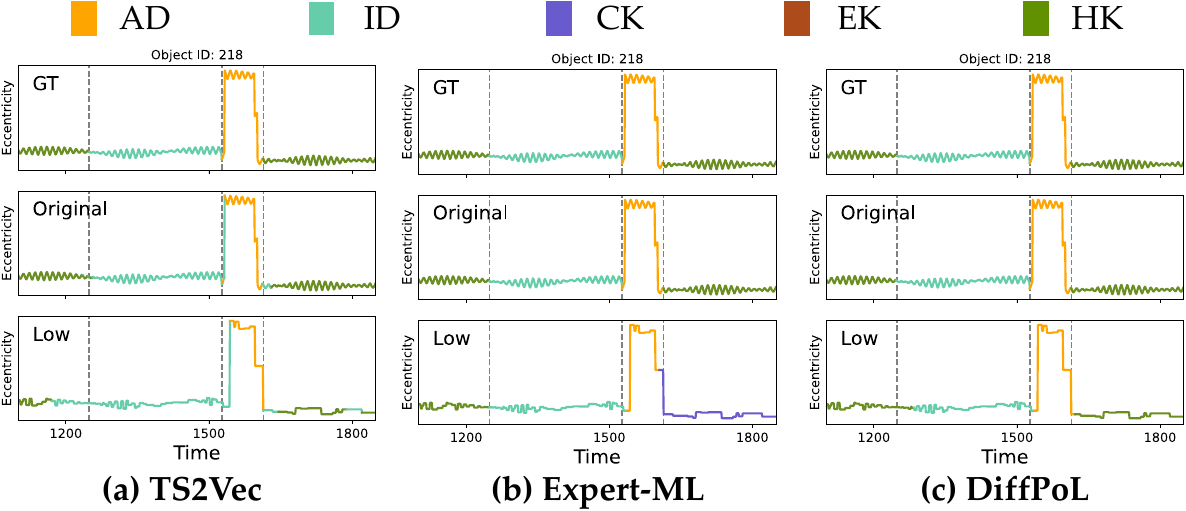}
    \caption{Visualization of identification results with different sampling rate data in EW direction. Top to bottom: Ground truth, results with original sampling rate, results with low sampling rate.
    }
    \label{fig:visual-low-eg}
\end{figure}

\subsection{Generality of \SystemName (RQ2)}\label{sec:exp-gen}

As illustrated in \cref{fig:sampling}, satellite positional data quality varies significantly between different ephemerides sources, with TLE-derived data providing substantially lower resolution compared to simulation tools and VCM-derived ephemerides. 
This discrepancy in data fidelity poses a significant challenge when deploying satellite PoL algorithms in real-world scenarios where only publicly available TLE data might be accessible. 
Thus, in this section, we further evaluate the utility of DiffPoL with satellite positional sequences derived from lower-resolution sources to assess its robustness under practical operational conditions.

\subsection{Ablation study (RQ3)}\label{sec:exp-abla}
\begin{table*}
    \small
    \centering
    \caption{Performance Comparison of Ablated Models in EW and NS directions. (Original / Low sampling rate)}
    \begin{tabular}{lcccccc}
    \toprule
    \multirow{2}{*}[\multirowoffset]{Ablated models} & \multicolumn{3}{c}{EW direction} & \multicolumn{3}{c}{NS direction} \\
    \cmidrule(lr){2-4} \cmidrule(lr){5-7}
    & F1 (\%) @ \{10\} ($\uparrow$) & Edit (\%) ($\uparrow$) & Acc (\%) ($\uparrow$) & F1 (\%) @ \{10\} ($\uparrow$) & Edit (\%) ($\uparrow$) & Acc (\%) ($\uparrow$) \\
    \midrule
    w/o decoder & 41.11 / 12.99 & 85.92 / 47.93 & \textbf{94.37} / 87.78 & 57.21 / 29.21 & 86.72 / 60.90 & 98.44 / 96.25 \\
    w/o geodetic & 87.30 / 75.70 & 87.53 / 80.38 & 91.58 / 83.08 & 94.69 / 89.69 & 95.84 / 88.53 & 98.62 / 94.92 \\
    w/o Sloss & 85.06 / 72.49 & 87.64 / 76.91 & 92.82 / 83.68 & 83.32 / 72.14 & 90.08 / 81.46 & 95.54 / 92.41 \\
    w/o mask & 91.09 / 78.07 & 91.61 / 83.51 & 93.14 / 85.14 & 92.72 / 86.48 & 95.56 / 88.07 & \textbf{98.97} / \textbf{96.60} \\
    DiffPoL & \textbf{91.68} / \textbf{83.15} & \textbf{93.00} / \textbf{86.41} & 94.10 / \textbf{88.91} & \textbf{94.79} / \textbf{92.69} & \textbf{96.72} / \textbf{93.11} & 98.20 / 96.59 \\
    \bottomrule
    \end{tabular}
    \label{tab:ablation-combined}
    
    \vspace{2pt}
    Bold indicates the best performance over ablated models. $\uparrow$: higher is better.
\end{table*}

\cref{fig:sampling-res-EW} shows the identification performance comparison across different methods under original and low sampling rates in EW direction. 
Specifically, the ``original sampling rate'' corresponds to the high-fidelity data with a uniform 2-hour interval, while the ``low sampling rate'' data was created by downsampling to an average interval of 13.7 hours (as shown in \cref{fig:sampling-pdf}) and then forward-filling to simulate realistic, sparse observations.

Specifically, while the performance of other methods, such as RF, drops noticeably under low sampling rates, \SystemName remains steady and robust. 
Expert-ML and \SystemName show similar high performance under the original sampling rate, with F1 scores close to 90. 
However, under the low sampling rate, Expert-ML's performance drops significantly, with F1 scores falling to around 60, while \SystemName maintains a relatively steady performance with F1 scores remaining above 80. 
The significant drop occurs because Expert-ML's rules depend on specific features that are destroyed by low sampling rates. 
For instance, a rule designed to detect a sharp step change will fail when downsampling and interpolation transform that step into a gradual change, rendering the maneuver invisible to the hard-coded logic.
In contrast, \SystemName's design allows it to handle the challenges posed by irregular sampling intervals effectively.
Similar trends were observed in the NS direction, confirming the method's robustness across different sampling rate.

We further select an object and visualize its identification results in EW direction with different sampling rate data in \cref{fig:visual-low-eg}.
TS2Vec, Expert-ML, and \SystemName are selected as the comparison.
The results of Expert-ML show considerable deviation from the ground truth, indicating its sensitivity to low sampling rates. 
In contrast, DiffPoL maintains consistent performance, closely matching the ground truth.
TS2Vec, due to its convolutional kernel-based design, shows stable performance in point-wise labeling but still exhibits fragmentation issues at state boundaries.
These observations highlight the effectiveness of \SystemName in handling positional sequences with low sampling rates, making it a more reliable choice for PoL identification for online operation scenarios.

To investigate the contribution of each sub-module to the overall performance of \SystemName, we carry out comprehensive experiments with the following ablated variants:

\begin{itemize}
    \item \textbf{w/o decoder:} 
    The decoder module is removed, only the encoder is trained with cross-entropy loss.    

    \item \textbf{w/o geodetic:} 
    The model takes only the six Keplerian elements as input, excluding the geodetic features, i.e., latitude, longitude, and altitude.

    \item \textbf{w/o Sloss:} 
    The segment-level losses introduced in \cref{sec:sub-training} are removed. 
    The model is trained only with the cross-entropy loss.

    \item \textbf{w/o mask:} The mask strategy mentioned in \cref{sec:sub-training} is removed. 

    
\end{itemize}

\cref{tab:ablation-combined} shows the performance comparison of the ablated models in EW and NS direction under both original and low sampling rates.
Results show that when removing the decoder, the model behaves similarly to other data-driven baselines and shows limited performance at the segment and sequential levels. 
This further underscores the crucial role of the denoising process in refining the sequence to achieve accurate identification.
For other ablated models, although the performance gap compared to \SystemName is not as significant as w/o decoder, there is still a noticeable decline, indicating that each sub-module contributes to the overall performance.
In particular, removing segment-level losses (w/o Sloss) leads to a significant drop in F1 and edit scores, highlighting the importance of the segment-level supervision mechanism in guiding the model to learn the sequential patterns effectively.
The mask strategy (w/o mask) shows a more noticeable impact on low sampling rates than the original sampling rate.
As EW maneuvers show patterns in longitude, the geodetic features (w/o geodetic) also play a crucial role in enhancing the model's performance.

\subsection{Computational cost (RQ4)}

\begin{table}
\centering
\caption{Comparison of computational costs across different methods.}
\label{tab:computational_cost}
\begin{tabular}{lccc}
\toprule
Method & Training time &  \begin{tabular}[c]{@{}l@{}}Inference time \\ (per satellite)\end{tabular} & \# of parameters \\
\midrule
Heuristic & - & 0.09 s & - \\
ClaSP & - & 0.81 s & - \\
RF & 5 minutes & 0.04 s & 3.06 M \\
CNN & 10 minutes & 0.01 s & 4.51 M \\
TS2Vec & 3.25 hours & 0.01 s & 1.81 M \\
Expert-ML & 12 minutes & 0.12 s & 5.71 M \\
\SystemName & 18 hours & 0.04 s & 1.90 M \\
\bottomrule
\end{tabular}
\end{table}

We evaluated the computational demands of \SystemName compared to baseline methods across three key metrics: training time, inference time, and model size. 
As shown in \cref{tab:computational_cost}, \SystemName requires a longer training period compared to traditional machine learning approaches like RF and Expert-ML.
This increased training time is expected given the complexity of diffusion models. 
However, \SystemName's inference speed is competitive with most baselines and significantly faster than methods like ClaSP  and Expert-ML. 
The efficient inference time is particularly important for practical deployment, as satellite positions are typically observed at intervals of hours rather than seconds, making \SystemName's processing speed more than sufficient for real-time satellite monitoring systems. 
This balance of model complexity and operational efficiency demonstrates that while \SystemName requires greater upfront training investment, its practical deployment costs remain reasonable for operational satellite tracking applications.

\subsection{Parameter sensitivity (RQ5)}\label{sec:exp-sens}

\begin{figure}[h]
    \centering
    \subfloat[Diffusion steps]{
        \includegraphics[width=0.9\linewidth]{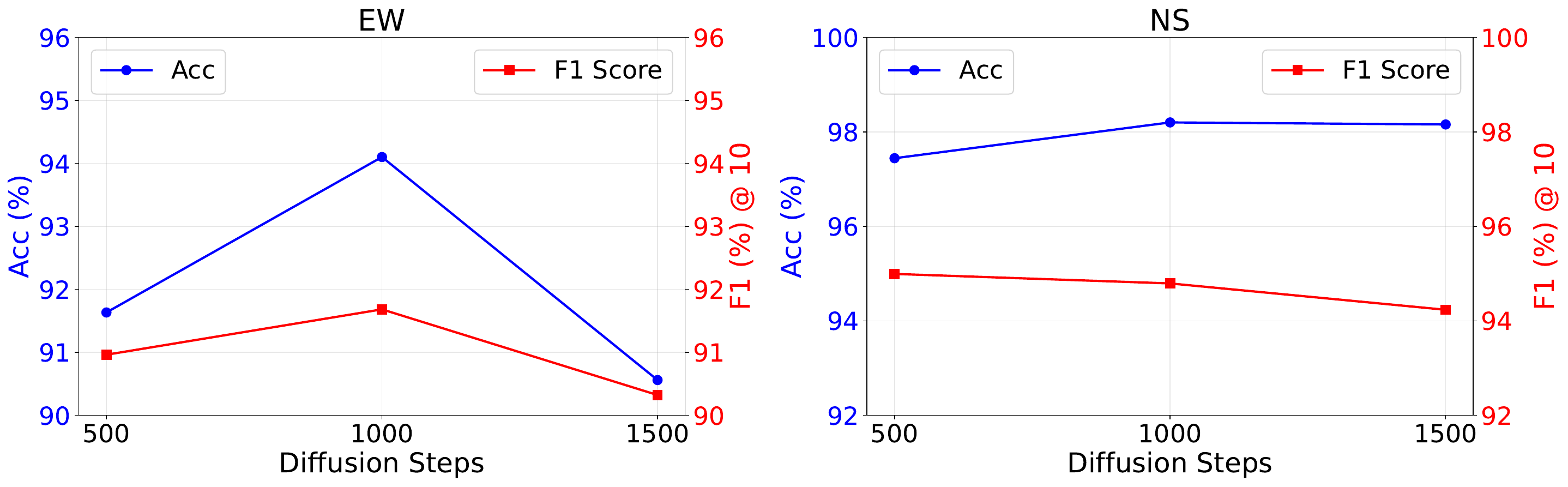}
        \label{fig:sensitivity-noise}
    }
    \\
    \subfloat[Sequence length]{
        \includegraphics[width=0.9\linewidth]{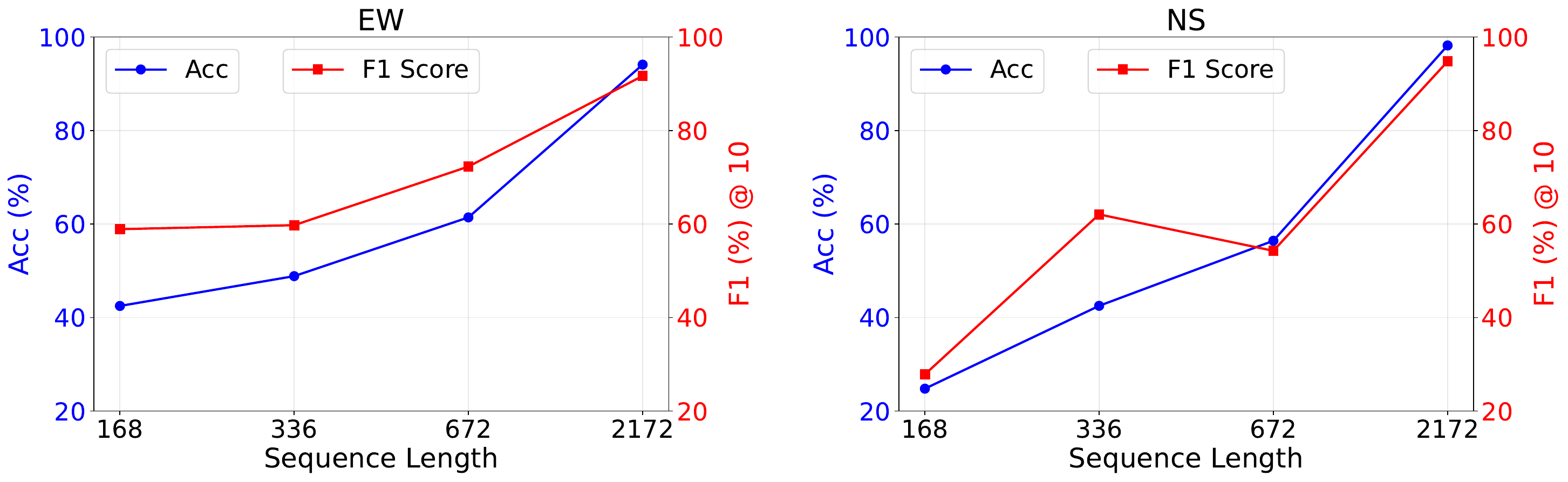}
        \label{fig:sensitivity-diffsteps}
    }
    \\
    \subfloat[Noise level]{
        \includegraphics[width=0.9\linewidth]{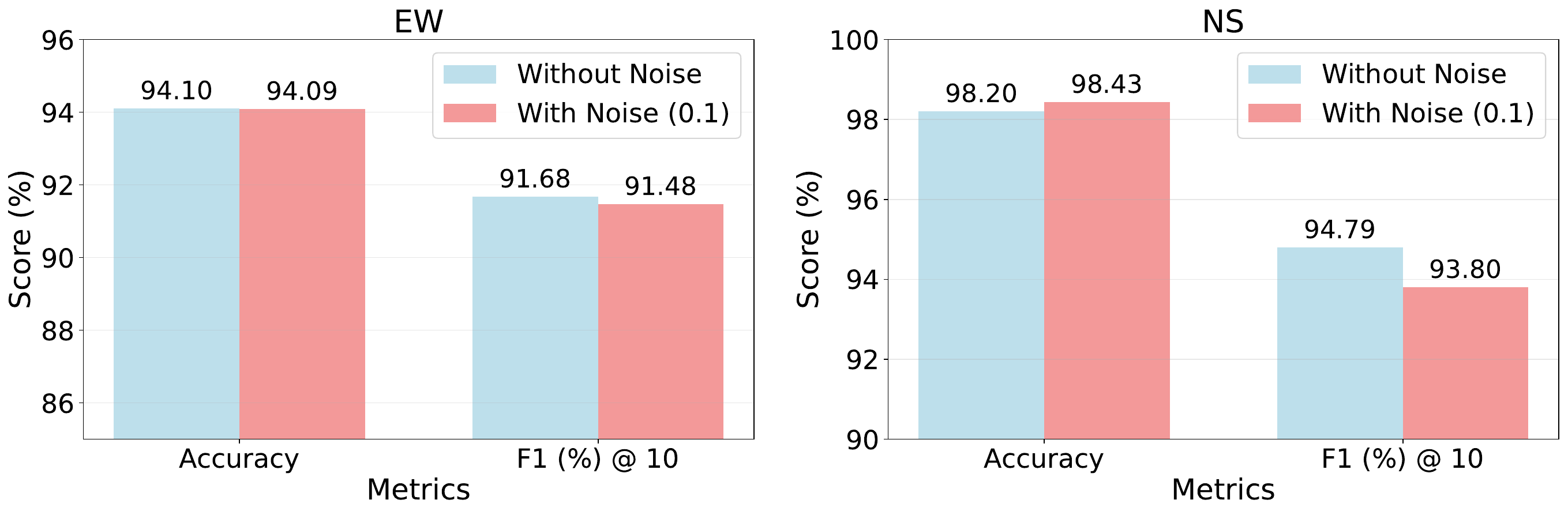}
        \label{fig:sensitivity-seqlen}
    }
    \caption{Performance sensitivity analysis of \SystemName with respect to  (a) number of diffusion steps, (b) sequence length, and (c) noise level}
    \label{fig:sensitivity}
\end{figure}

To evaluate the sensitivity of \SystemName to its key hyperparameters, we conduct a series of experiments by varying one parameter at a time while keeping others at their default values. We investigate the impact of three critical parameters: the number of diffusion steps, the input sequence length, and the level of input noise. 
i) For the diffusion steps, we vary the total number of steps from 500 to 1500 to analyze its effect on the denoising process. 
ii) To assess how the model handles different temporal contexts, we experiment with various input sequence lengths, ranging from two weeks (168 steps) to two months (672 steps). 
iii) Finally, to simulate real-world measurement errors and test the model's robustness, we add zero-mean Gaussian noise with standard deviations at 0.1 to the normalized input data. 
The results of these sensitivity analyses are presented in \cref{fig:sensitivity}.

As shown in \cref{fig:sensitivity-noise}, the model's performance is slightly sensitive to the number of diffusion steps. 
The default setting of 1000 steps yields the optimal results. 
A lower number of steps leads to performance degradation, likely due to an insufficient denoising process that fails to fully refine the initial predictions. 
Conversely, increasing the steps beyond the optimal point also results in a slight performance drop, which may be attributed to overfitting.
\cref{fig:sensitivity-diffsteps} shows that performance drops significantly as the sequence length decreases, with the effect being more pronounced in the NS direction. 
This phenomenon is linked to the temporal patterns of PoL maneuvers. 
PoLs in the EW direction are often correlated with daily periodicity, which can be captured in shorter time windows, while NS maneuvers are frequently related to longer-term cycles like lunar periodicity, requiring a more extended context. 
Moreover, since adjacent PoL states are correlated, reducing the sequence length leads to a loss of this contextual information, further degrading performance.
Finally, \cref{fig:sensitivity-seqlen} illustrates that the model's performance remains remarkably stable with increasing levels of input noise, demonstrating its robustness to measurement errors, which is a critical attribute for real-world applications.

\begin{figure*}
    \centering
    \includegraphics[width=0.8\linewidth]{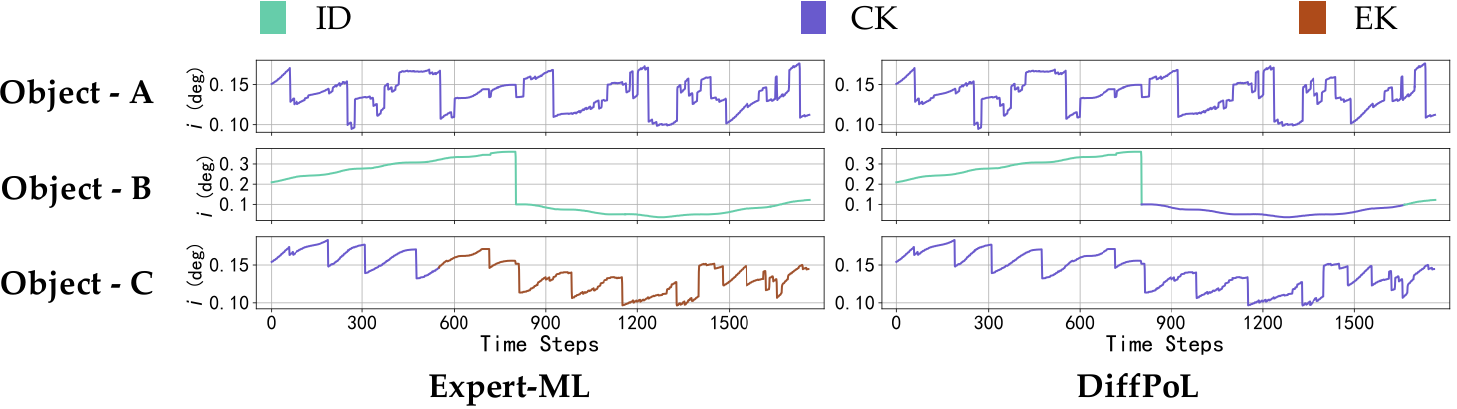}
    \caption{Visualization of identification results on real-world GEO satellites in NS direction.}
    \label{fig:result_NWT}
\end{figure*}

\section{Case Study on Real-world Data}
\label{sec:case_study}

In this section, we present a case study using real-world satellite positional data to demonstrate the effectiveness of our proposed method. The data include orbital data of multiple GEO satellites in operation over six months. The data is resampled to 2-hour intervals to align with the settings in the simulation study. As satellites are operated by multiple entities, the PoL labels are not available. Due to confidentiality requirements, we do not disclose the exact satellite names and timestamps.

\cref{fig:result_NWT} visualizes the identification results of Expert-ML (left column) and proposed \SystemName (right column) on three representative GEO satellites, labeled as Object-A, Object-B, and Object-C. 
The vertical axis represents inclination values in degrees, while the horizontal axis shows time progression in steps. 
The results highlight both similarities and differences between the two methodologies. 
For Object-A, both methods demonstrate consistent pattern recognition in the inclination data, showing strong agreement in operational mode identification. 
For Object-B, \SystemName successfully identifies the critical transition where the inclination pattern changes significantly, whereas Expert-ML fails to detect this important shift. 
This discrepancy likely stems from Expert-ML's dependence on predefined expert labels, which may not accurately capture the actual operational phases of the satellite. 
For Object-C, \SystemName provides a more coherent and reasonable identification pattern, as it recognizes that the station-keeping proportion type for a specific satellite typically maintains consistency throughout its operational period rather than frequently alternating between different modes.

These results validate that \SystemName can effectively discover meaningful PoL patterns in real-world GEO satellite data without relying on expert-defined labels, demonstrating its practical utility for space situational awareness applications.
\section{Conclusion}\label{sec:conclusion}


This paper presents a novel data-driven approach to satellite pattern-of-life (PoL) identification, \SystemName, marking a significant shift from traditional expert-dependent methods toward autonomous solutions. 
Our method addresses the challenges posed by the complexity and variability of aerospace systems by learning directly from orbital elements and positional data rather than relying on manually crafted rules. 
By utilizing a multivariate time-series encoder and a tailored decoder within a diffusion model, \SystemName achieves end-to-end PoL identification in a sequence PoL label generation manner, which is the first exploration of data-driven end-to-end PoL identification with diffusion models.

Experiments on satellite datasets prove \SystemName can identify satellite PoL in an end-to-end manner without the need for manual refinement or domain-specific feature engineering. 
Comparison over state-of-the-art methods demonstrates the superiority of \SystemName, especially in sequence-level evaluation.
Additionally, \SystemName exhibits robust performance when processing data from different ephemerides sources with varying levels of precision and resolution, making it significantly more applicable to real-world scenarios. 
This novel approach not only enhances the scalability and applicability of PoL identification but also contributes to improved space situational awareness and satellite monitoring. 
For future work, we aim to explore the application of \SystemName on a broader range of satellites and aerospace systems.

\bibliographystyle{IEEEtran}
\bibliography{ref}

\begin{IEEEbiography}[{\includegraphics[width=1in,height=1.25in,clip,keepaspectratio]{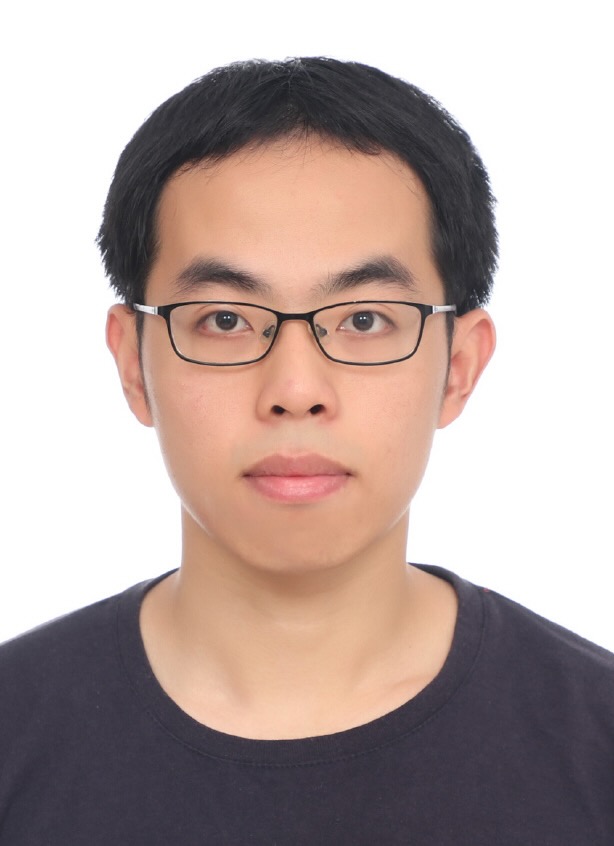}}]{Yongchao Ye }(Student Member, IEEE)
is currently a Ph.D. student with Department of Data Science, College of Computing, City University of Hong Kong. He received the MPhil degree from Department of Computer Science and Engineering, Southern University of Science and Technology, Shenzhen, China, in 2023 and the B.Eng degree in computer science and technology from Ningbo University, Ningbo, China, in 2020. His research interests include spatio-temporal data mining, deep generative model, and their applications in aerospace systems and intelligent transportation systems.
\end{IEEEbiography}

\begin{IEEEbiography}
[{\includegraphics[width=1in,height=1.25in,clip,keepaspectratio]{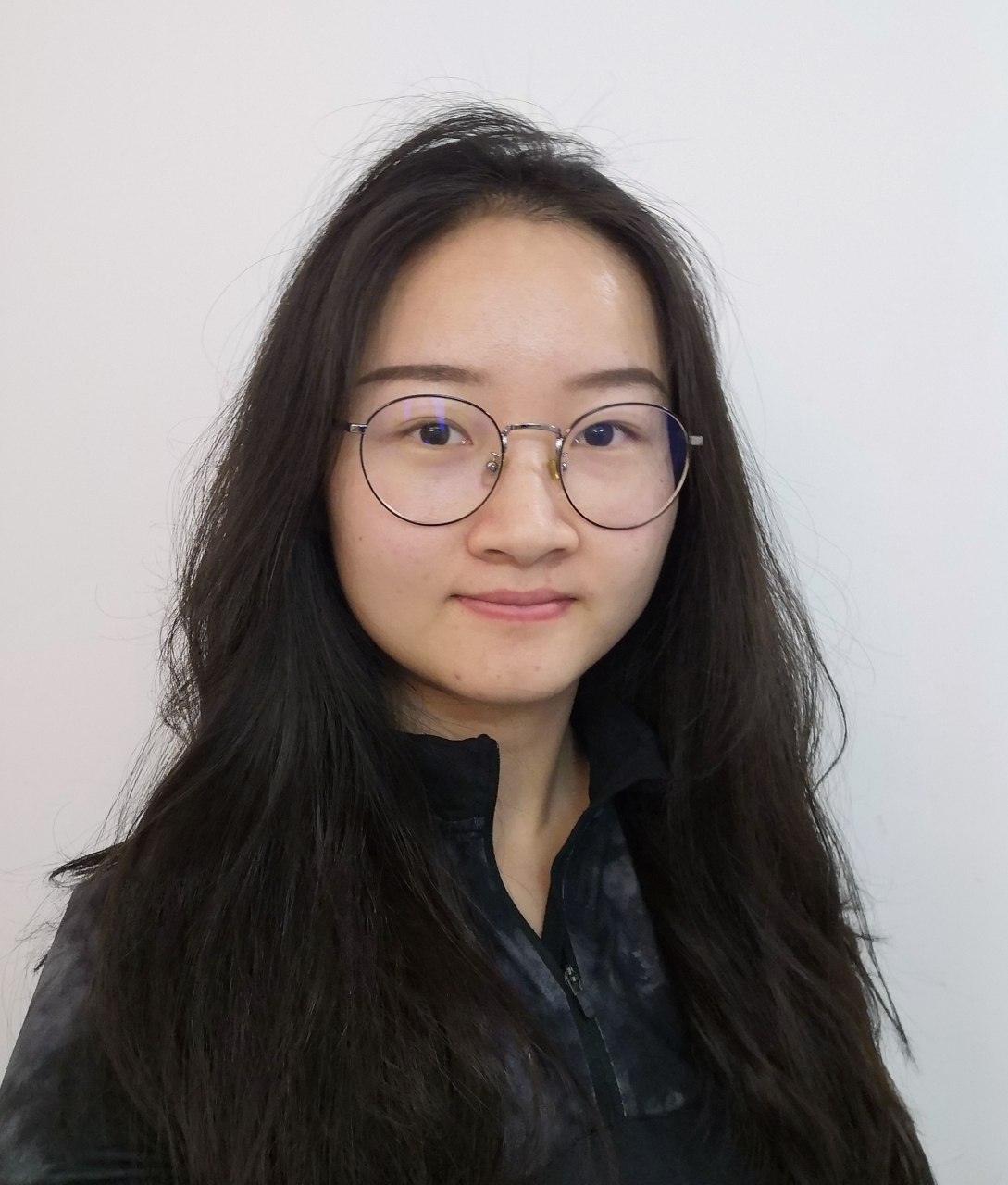}}]{Xinting Zhu }(Member, IEEE) received her B.Eng. degree in Aircraft Airworthiness Engineering from the Department of Transportation Science and Technology, Beihang University, and her Ph.D. degree in Data Science from the Department of Data Science, City University of Hong Kong. She is currently a Jockey Club Global STEM Postdoctoral Fellow at City University of Hong Kong, supported by The Hong Kong Jockey Club Charities Trust.
Her research interests include data analytics, machine learning, deep learning, and their applications to aerospace operations and management.
\end{IEEEbiography}

\begin{IEEEbiography}
[{\includegraphics[width=1in,height=1.25in,clip,keepaspectratio]{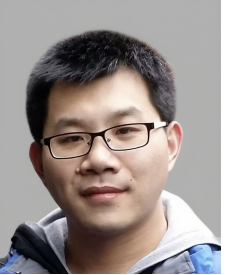}}]
{Xuejin Shen} is currently employed by Chengdu Atom Data Technology Co., Ltd. He obtained his B.S. and M.S. degrees in Computer Science and Technology from Taiyuan University of Technology. His research interests include data analysis, machine learning, deep learning, and their applications to aerospace operations and aircraft test flight.
\end{IEEEbiography}

\begin{IEEEbiography}
[{\includegraphics[width=1in,height=1.25in,clip,keepaspectratio]{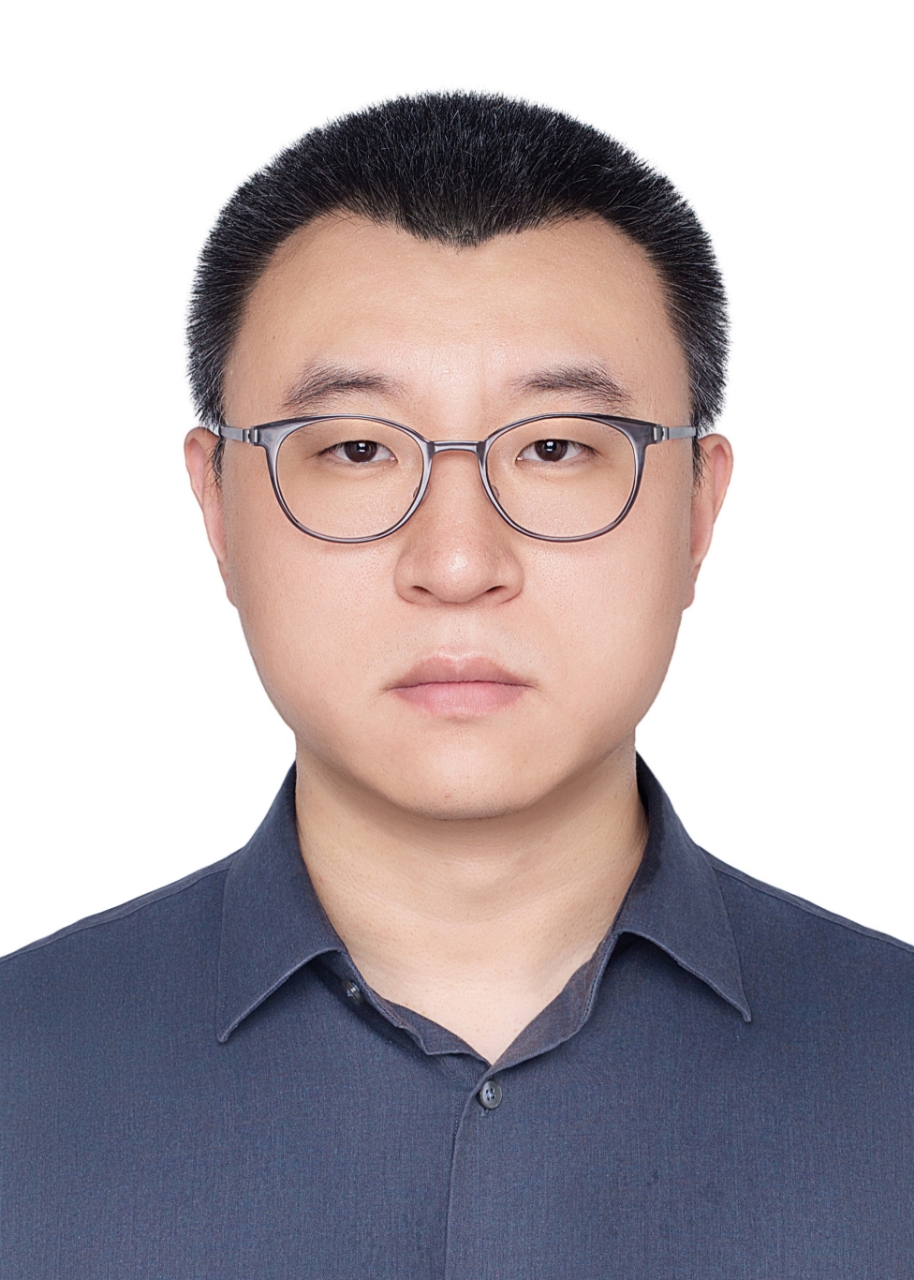}}]
{Xiaoyu Chen} is the CEO \& CTO of Chengdu Atom Data Technology Co., Ltd. He obtained his bachelor's degree in Automation from South China University of Technology. His research interests include data analysis, deep learning, and their applications to aerospace operations and aircraft test flight.
\end{IEEEbiography}

\begin{IEEEbiography}
[{\includegraphics[width=1in,height=1.25in,clip,keepaspectratio]{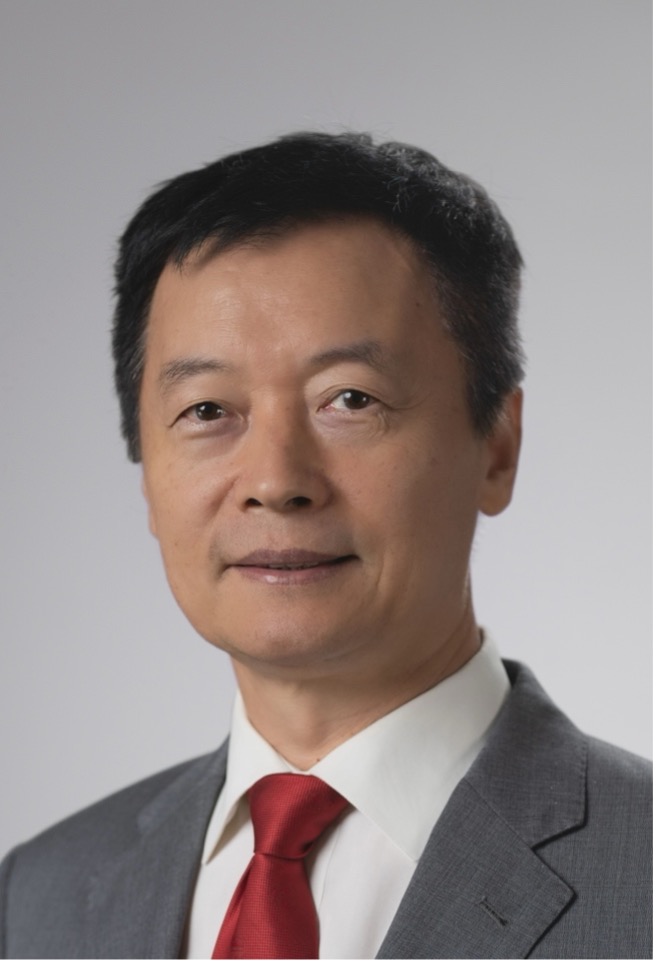}}]
{S. Joe Qin }(Fellow, IEEE) is the Wai Kee Kau Chair Professor of Data and President of Lingnan University in Hong Kong. He obtained his B.S. and M.S. degrees in Automatic Control from Tsinghua University in Beijing and his Ph.D. degree in Chemical Engineering from University of Maryland at College Park. Qin’s research interests include data science and analytics, statistical and machine learning, industrial AI, process monitoring, model predictive control, system identification, smart manufacturing, and smart energy management. Dr. Qin is a member of the European Academy of Sciences and Arts and Fellow of the Hong Kong Academy of Engineering, the U.S. National Academy of Inventors, IFAC, AIChE, and IEEE. He is the recipient of the 2022 AIChE CAST Computing Award, 2022 IEEE CSS Transition to Practice Award, the U.S. NSF CAREER Award. 
\end{IEEEbiography}

\begin{IEEEbiography}
[{\includegraphics[width=1in,height=1.25in,clip,keepaspectratio]{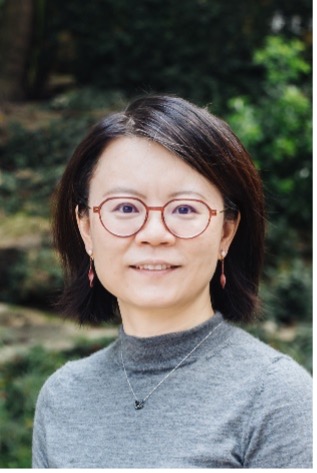}}]
{Lishuai Li} (Senior Member, IEEE) received her Ph.D. and M.Sc. degrees from the Department of Aeronautics and Astronautics at Massachusetts Institute of Technology (MIT), Cambridge, MA, USA, and B.Eng. degree in aircraft design and engineering from Fudan University, Shanghai, China. She is a Professor with the Dept. of Data Science, Dept. of Systems Engineering, and Dept. of Mechanical Engineering at City University of Hong Kong, Hong Kong, China. She develops analytical methods—from statistical approaches to Artificial Intelligence—for operational excellence of complex systems, e.g. industrial intelligence, low altitude economy, and air transportation systems.
\end{IEEEbiography}

\end{document}